\documentclass[lettersize,journal]{IEEEtran}
\usepackage{amsmath,amsfonts}
\usepackage{array}
\usepackage[caption=false,font=footnotesize,labelfont=sf,textfont=sf]{subfig}
\usepackage{textcomp}
\usepackage{stfloats}
\usepackage{url}
\usepackage{verbatim}
\usepackage{graphicx}
\usepackage{cite}
\usepackage{xcolor}

\usepackage{float}
\usepackage{booktabs}
\usepackage{url}
\usepackage{multirow}
\usepackage{xspace}
\usepackage{listings}
\usepackage{amsthm}
\usepackage{enumerate}
\usepackage{enumitem}
\usepackage{footmisc}
\usepackage[figuresright]{rotating}

\usepackage[ruled,noline,linesnumbered,noend]{algorithm2e}

\newtheorem{definition}{Definition}
\newtheorem{theorem}{Theorem}
\newcommand{\Paragraph}[1]{~\vspace*{-0.9\baselineskip}\\{\bf #1}}
\newcommand{\Figure}{Fig.\xspace}
\newcommand{\Equation}{Equ.\xspace}

\begin{document}

\title{gBuilder: A Scalable Knowledge Graph Construction System for Unstructured Corpus}

\author{Yanzeng~Li~and~Lei~Zou
 \thanks{Y. Li is with Wangxuan Institute of Computer Technology, Peking University, Beijing 100871, China.
 (e-mail: liyanzeng@stu.pku.edu.cn).}
 \thanks{L. Zou is with Wangxuan Institute of Computer Technology, Peking University, Beijing 100871, China.
 (e-mail: zoulei@pku.edu.cn).}
 \thanks{\textit{(Corresponding author: Lei Zou)}}
}



\maketitle

\begin{abstract}
We develop gBuilder, a scalable and user-friendly knowledge graph construction (KGC) system that extracts structured knowledge from unstructured textual corpora. Different from existing KGC systems, gBuilder features a flexible and user-defined pipeline to accommodate the rapid development of information extraction (IE) models. The system offers a wide selection of built-in IE operators, including heuristic-based, deep learning-based and even large language model-based extraction methods, as well as custom programmable operators, that can be adapted to data from different domains. 
To enable practical large-scale deployment capable of serving multiple users and processing comprehensive KGC, gBuilder runs on the cloud and we design a cloud-based, self-adaptive task scheduling algorithm for it to ensure scalability and economical efficiency throughout complicated IE pipelines. 
Experimental evaluations confirm gBuilder's capacity to organize diverse IE models to extract reliable structured information, while demonstrating its scalability and cost-efficiency for large-scale KGC tasks. 
\end{abstract}


\section{Introduction}

\IEEEPARstart{R}ecently, knowledge graph has attracted significant attentions in several research communities, including database \cite{Zou2013gStoreAG,wilkinson2006jena,zhang2015deepdive}, natural language processing (NLP) \cite{schneider2022decade, zou2020survey, wang2017knowledge} and machine learning \cite{nickel2015review,tiddi2022knowledge}.
Even in the era of Large Language Models (LLMs), knowledge graph also plays some important role, which can be considered as reliable and interpretable sources of information or symbolic memories, thereby enriching the generated content and enhancing the capabilities of LLMs.~\cite{wang2023knowledgpt, feng2023trends, zhan2023admus}. 
Therefore, the demand for knowledge graph resources has promoted research on knowledge graph construction (KGC) techniques. For example, DBpedia\cite{auer2007dbpedia}, Freebase\cite{bollacker2008freebase} and YAGO\cite{hoffart2013yago2} build large-scale open-domain knowledge graph datasets, while, for specific domain applications~\cite{abu2021domain} such as medical, agricultural, and financial, domain-specific knowledge bases are required, placing significant demands on KGC.

Actually, KGC has a long and tedious workflow to extract knowledge from unstructured data and represent them as a knowledge graph, including schema design\cite{kondylakis2009ontology, noy1997state}, entity extraction~\cite{nadeau2007survey, li2020survey} and relation extraction\cite{kumar2017survey} and so on. Although there is a lot of research work related to KGC and a bunch of toolkit are also available, they only focus on some specific steps, such as proposing an relation extraction model~\cite{kumar2017survey} or ontology merge strategy \cite{de2006ontology,kondylakis2009ontology,stumme2001ontology}. 
Traditionally, building a knowledge graph is based on manual annotation or expert systems~\cite{carley1988formalizing,kondreddi2014combining,sarasua2015crowdsourcing}, but it is often laborious, expensive~\cite{paulheim2018much} and only adapted to the small-scale corpus.
Therefore, researchers adopt some rule-based methods~\cite{auer2007dbpedia,bollacker2008freebase,hoffart2013yago2} to obtain structured knowledge from un/semi-structured text to build knowledge graphs.
Benefiting from the development of machine- and deep-learning techniques, much NLP research on information extraction (IE) has emerged. Deep learning-based models can be trained by supervised datasets or other paradigms to extract entities or relations from text with high accuracy~\cite{Fensel2020}.
In particular, pre-trained language models (PLMs) have taken the performance of IE models to a new level~\cite{BERT, GPT, XLNet}, providing a solid foundation for extracting knowledge from unstructured corpora. 
Moreover, recent advancements in LLMs like ChatGPT~\cite{openai2023gpt4} and Llama~\cite{touvron2023llama} further empowers IE by prompting models to generate desired information in suitable format, guided by a set of demonstration samples~\cite{wei2023zero, gupta2022matscibert, zhang2023llmaaa}.

However, in practice, domain experts often lack the knowledge of complicated IE models and the coding capability to connect these models in a pipeline fashion to accomplish a KGC task. Therefore, designing KGC’s pipeline and organizing various IE models has become a new problem. To the best of our knowledge, only a few of systematic proposals exist, e.g., DeepDive~\cite{zhang2015deepdive} DeepKE~\cite{arxiv.2201.03335} and T2KG~\cite{kertkeidkachorn2017t2kg}. However, these KGC systems have the following issues:
(1) They only embed simple IE models, such as heuristic rules and probabilistic statistical models to extract entities and relations. Their fixed paradigms cannot utilize state-of-the-art deep learning models~\cite{li-etal-2021-tdeer, sui2020joint, wang2022deepstruct}.
(2) Users are restricted to the KGC systems' paradigms with little flexibility or operating space. 
(3) Existing KGC systems do not consider scalability and performance of the inference process, inefficient for large-scale KGC tasks.
(4) LLM-based KGC pipelines can only utilize a limited number of demonstration samples, leading to the underutilization of the labeled corpus. Additionally, they suffer from extreme slowness and high costs.

Generally, a desirable KGC system should have the following features, motivating our proposed system, \texttt{gBuilder}:
(1) \textbf{Flexible:} The KGC system should embrace the rapid development of IE models instead of a fixed build-in models, meanwhile, the KGC pipeline should adapt to different scenarios rather than a pre-defined paradigm.
(2) \textbf{User-friendly:} An easy-to-use interface should allow domain experts to design KGC tasks without knowledge of complicated IE models or coding from scratch.
(3) \textbf{Efficient \& Cost-efficiently:} The system's efficiency, scalability and cost-efficiency also should be considered in KGC, which is important for large-scale KGC progress. 

Therefore, \texttt{gBuilder} system considers all of the above features. First, we abstract NLP models and operators from the systematic perspective and design a unified KGC framework that can involve any IE model flexibly. 
In gBuilder, ordinary users can add any off-the-shelf IE model, while expert users can also import custom trained models. Furthermore, we abstract IE models and built-in operators for uniting these models, based on which, \texttt{Flowline}, a combination of IE workflow and data pipeline, is proposed to design a KGC task. Note that the \texttt{Flowline} design varies on different KGC tasks, not a fixed paradigm. 
Second, \texttt{gBuilder} is friendly to ordinary users by providing a visual drag $\&$ drop design interface to support domain experts to describe a \texttt{Flowline} with no/low code manner. 
Finally, to ensure the scalability of KGC and provide stable and efficient service for multiple users, \texttt{gBuilder} adopts cloud distribution architecture, proposes a series of efficient task decomposition and resource scheduling strategies to boost the efficiency. Specifically, we propose a DAG-based \texttt{Flowline} partitioning method to parallel KGC tasks and design cloud-based self-adaptive task scheduler according to the characteristics of deep learning-based IE tasks, which enables scalability and cost-efficiency on large-scale KGC. 

We summarize and highlight the contributions of our proposed system in the following points: 
\begin{enumerate}[leftmargin=*,align=left]
    \item We investigate the practical process of KGC and propose \texttt{gBuilder}, a system for flexibly building KGC workflow and data pipelines, achieving the goal of constructing the knowledge graph with low cost and little background knowledge for ordinary users; 
    \item We investigate plenty of deep learning-based IE models and ontology merging theory, and normalize diverse IE models into three categories, provide a series of template-based, heuristic-based and programmable operators for adapting built-in models and data from different domains. 
    \item We adopt the cloud-based distributed solution and design a cloud-based self-adaptive task scheduler according to the characteristics of IE tasks, which enables scalability on large-scale KGC. 
    \item Experimental results on benchmark datasets demonstrate that superiority of \texttt{gBuilder} in the effectiveness of KGC tasks compared to existing KGC systems. Furthermore, real-world scheduling experiments prove our DAG-based task partitioning and cloud-based resource scheduling strategies outperforms state-of-art counterparts in both scalability and economical efficiency. 
 \end{enumerate}
In addition, we design a user-friendly drag-and-drop interactive interface for designing the KGC process, as well as provided operation interfaces such as data management and model management. 

\section{Preliminary \& Related Work}\label{sec:pre}
 
Knowledge graph construction (KGC) studies have emerged in recent years, including the ontology design for various domain-specific knowledge graphs~\cite{abu2021domain, kejriwal2019domain}, the KGC by top-down~\cite{hoffart2013yago2} or bottom-up~\cite{auer2007dbpedia} approaches, and the utilization of NLP methods such as deep learning-based models to extract structured information from unstructured corpus~\cite{zou2020survey, wang2018information}. In this paper, based on abundant IE models, we design \texttt{gBuilder}, a KGC system focusing on choreographing the models and operators to implement the whole KGC task. Before digging into details, we first introduce preliminary knowledge and related work. 

\Paragraph{Knowledge Graph (KG)}. Knowledge Graph~\cite{Fensel2020} uses a graph-structured data model to store and govern linked data and can explicitly describe the semantics for the connection of structured data. Typically, Resource Description Framework (RDF) is the de-facto standard of KG, representing KG as a collection of triples (\textit{subject}, \textit{predicate}, \textit{object}) (abbreviated as $(s,p,o)$), which defines the relation between two entities ($s$ and $o$) or describe the attribute of an entity ($s$).

\Paragraph{Ontology}. An ontology defines the data patterns and characterizes the high-level conceptualization of the formal semantics of a KG, which can be regarded as the \emph{schema} of the data. With the guidance and restriction of a well-defined ontology, the process of KGC would be normalized, and the constructed product would be consistent. Specifically, the ontology specifies the target entities, relations, and attributes to extract~\cite{anantharangachar2013ontology}. In applications, experts can construct ontology manually, and many ontology design tools are available, including Protégé~\cite{musen2015protege}. In this paper, we focus on the information extraction phase of KGC and assume that an ontology has been explicitly defined. 

\begin{figure}[t]
\centering{\includegraphics[width=.9\linewidth]{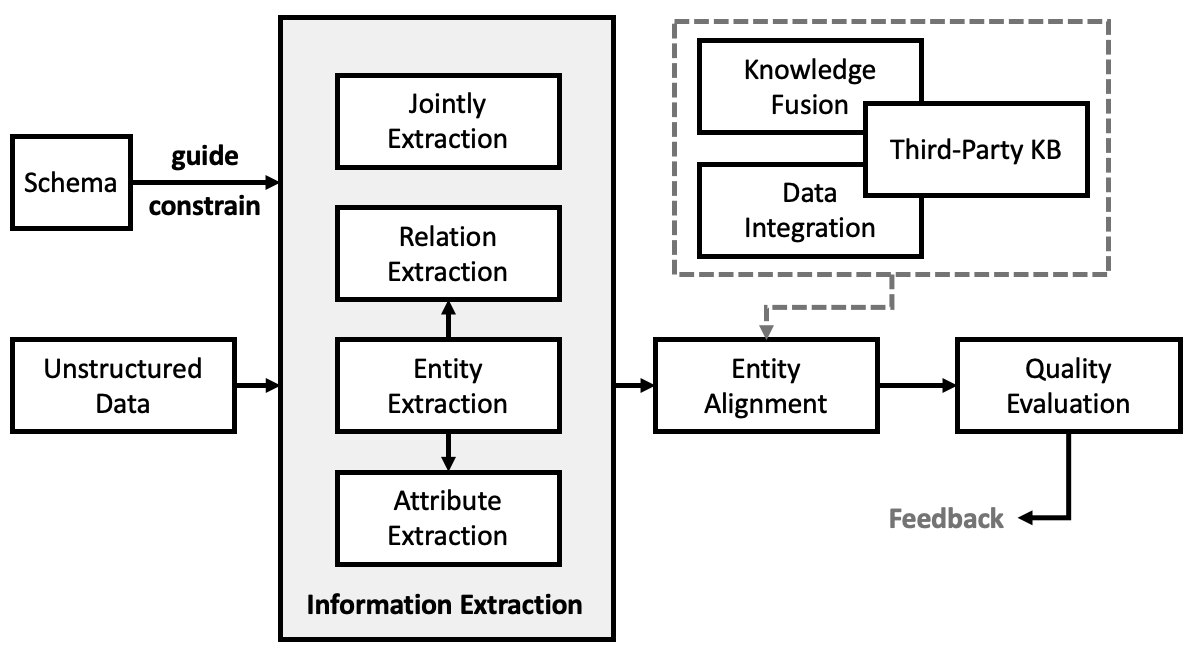}}
	\caption{The illustration of top-down KGC from unstructured text. gBuilder mainly concerns the grey highlighted information extraction part.}
	\label{fig:KGC}
	\vspace{-0.5cm}
\end{figure}
\Paragraph{Knowledge Graph Construction (KGC)}. Building a KG from an unstructured corpus is a critical technology path to building up a more complete and larger-scale knowledge graph \cite{chen2020review, zhao2018architecture}. 
KGC is generally categorized into two approaches: top-down and bottom-up. The top-down approach means the target ontology is well-defined, and then knowledge instances obeying the ontology are appended into the knowledge base \cite{zhao2018architecture}, while, the bottom-up scheme extracts knowledge instances directly, and the ontologies are summarized from populated instances~\cite{lee2007automated}. gBuilder belongs to the former one. \Figure \ref{fig:KGC} demonstrates the mainstream top-down KGC process. Note that the whole KGC lifecycle should also include knowledge fusion and KB (knowledge base) quality control, but these are beyond the scope of this paper. We focus on the IE (information extraction) component in this paper, which extracts the knowledge instances under the guidance of the target ontology. 
Generally, IE is an NLP task that involves extracting structured data from unstructured or semi-structured text. In order to automatically acquire knowledge instances, including entities, relations, and attributes, numerous IE tasks are extensively studied: 
\begin{figure}[t]
\centering{\includegraphics[width=\linewidth]{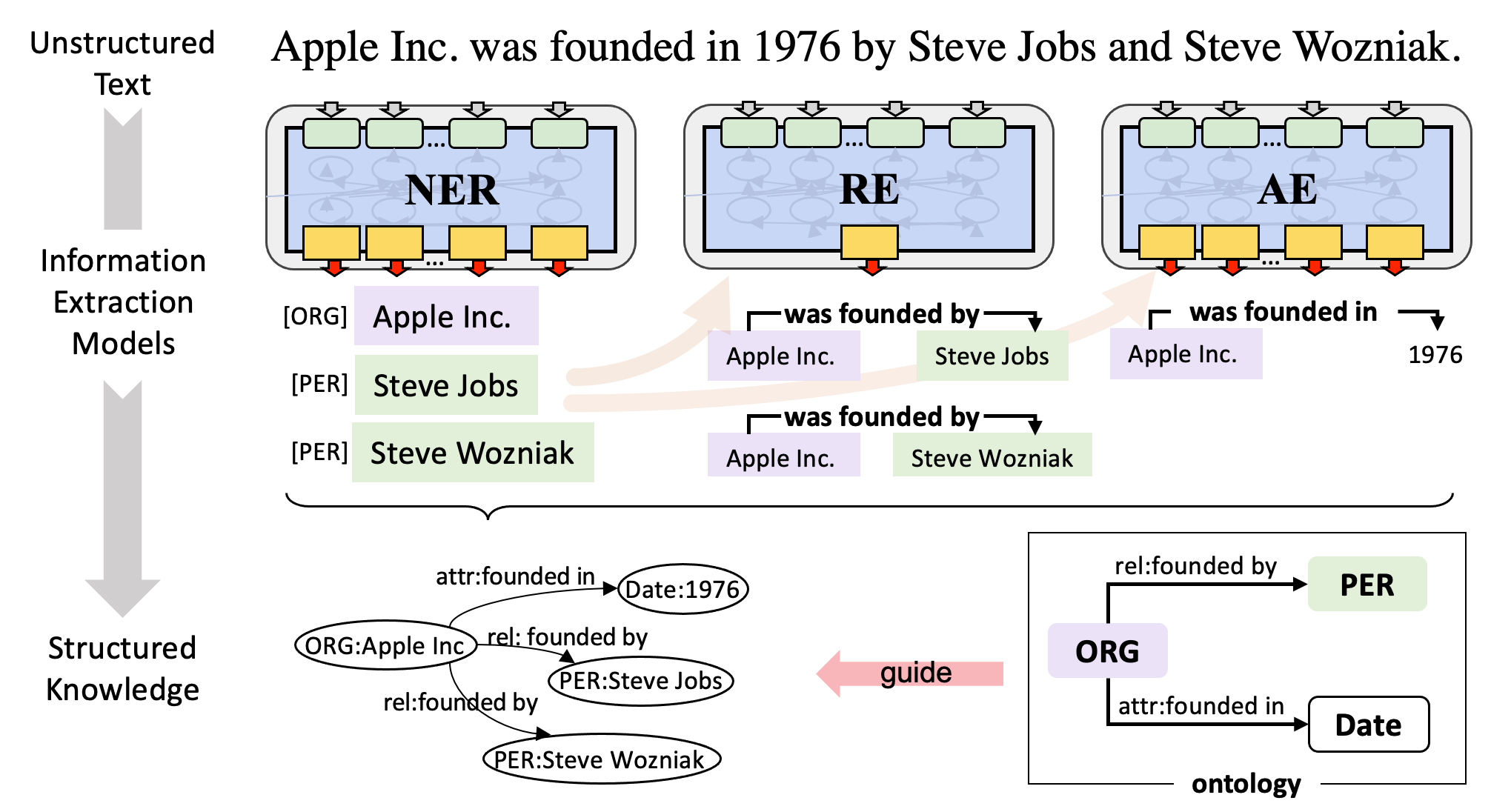}}
	\caption{A schematic of gBuilder KGC with multiple deep learning IE models guided by pre-defined ontology.}
	\label{fig:kgc-exp}
	\vspace{-0.3cm}
\end{figure}

\Paragraph{Named Entity Recognition (NER)} aims to extract mentions of rigid designators from text belonging to pre-defined semantic types such as Person, Location, Organization \cite{nadeau2007survey, li2020survey}. The entity is the essential node in the knowledge graph, and the quality of entity extraction (precision and recall) greatly impacts the quality of follow-up tasks for knowledge acquisition. 

\Paragraph{Relation Extraction (RE)} refers to identify the relationship for entity pairs. In the pre-defined schema scenario, it is usually treated as a classification task to predict the relation category according to evidence text containing mentions of the entity pair~\cite{kumar2017survey}. The triples constituted by relations are the fundamental instance in the knowledge graph, and the performance of RE is directly related to the accuracy of edges and connectivity of the knowledge graph~\cite{10.1145/3583780.3615178}. 

\Paragraph{Jointly Extraction (JE)} is an emerging IE approach in recent years, and it aims to detect entity pairs jointly along with their relations by a single model. JE is treated as a supervised learning task based on structured annotation, and it usually gains better performance on specific datasets due to more supervisory signals, and JE eliminates cascading errors in the NER-RE pipeline utilizing the superiority of the joint monomer model \cite{wei2019novel, soares2019matching}. In IE for KGC, there is a trade-off between JE's high performance and its expensive annotation cost and rigid output structure~\cite{sun2019distantly}.

\Paragraph{Attribute Extraction (AE)} is a task to extract the literal attribute and the corresponding attribute value for a specific entity from unstructured text. AE can be divided into attribute value extraction and name-value joint attribute extraction~\cite{zhang2021semi, li-etal-2023-attgen}. The former can be solved by the regex matching or the neural network similar to NER, and the latter can be solved simultaneously with RE and JE \cite{wei2019novel}, or designed heuristic rule \cite{vandic2012faceted, more2016attribute}. 

As illustrated in \Figure \ref{fig:kgc-exp}, the above IE models work cooperatively to accomplish their respective duties, and structured knowledge can be extracted from unstructured text.

\Paragraph{Challenges of IE tasks in KGC}: In recent years, the NLP community has put forth a wide variety of IE models, claiming to achieve state-of-the-art~(SOTA) on various datasets. 
A up-to-date approach is to leverage the power of LLMs by using a concise prompt that describes the task, verbalizes the ontology classes, and instructs the model on the desired output format~\cite{zhang2023llmaaa, wang2023gptner}\footnote{Appendix~\ref{app:prompts} demonstrates typical prompts used for extracting entities and relations.}.
The existence and forthcoming of different IE tasks and model forms brings significant diversity, making it challenging to effectively organize KGC efforts while simultaneously harnessing the unique strengths of different models.

The diversity of IE tasks and dazzling array of NLP models bring uncertainty and pose challenges to KGC, especially when domain engineers lack of Machine Learning (ML) and NLP experience, which is the bottleneck of KGC in real applications. Existing KGC systems~\cite{zhang2015deepdive,arxiv.2201.03335} perform KGC processes such as relation/attribute extraction under a fixed paradigm and do not take advantage of the latest SOTA IE models. For example, DeepDive~\cite{zhang2015deepdive} proposes heuristic and rule-based extraction but fails to employ SOTA NLP models. Even the latest work, DeepKE~\cite{arxiv.2201.03335}, which introduces deep learning models for KGC in various scenarios, is still under a fixed KGC paradigm and cannot utilize the inexhaustible IE models for building the whole KGC workflow.

To enable the KGC process in a manufacturer model, we should develop a unified KGC paradigm in gBuilder to address the following technical challenges of IE tasks in KGC: (1) how to build a detachable and replaceable system to embrace the rapid development of IE models in the NLP community rather than with fixed models; (2) how to abstract the diverse IE tasks and models in a normalized form and how to design a low-code interface for KGC in a friendly and flexible way; (3) last but not least, the efficient processing of IE tasks is also vital in KGC pipeline. 

\begin{figure}[h]
\centering{\includegraphics[width=\linewidth]{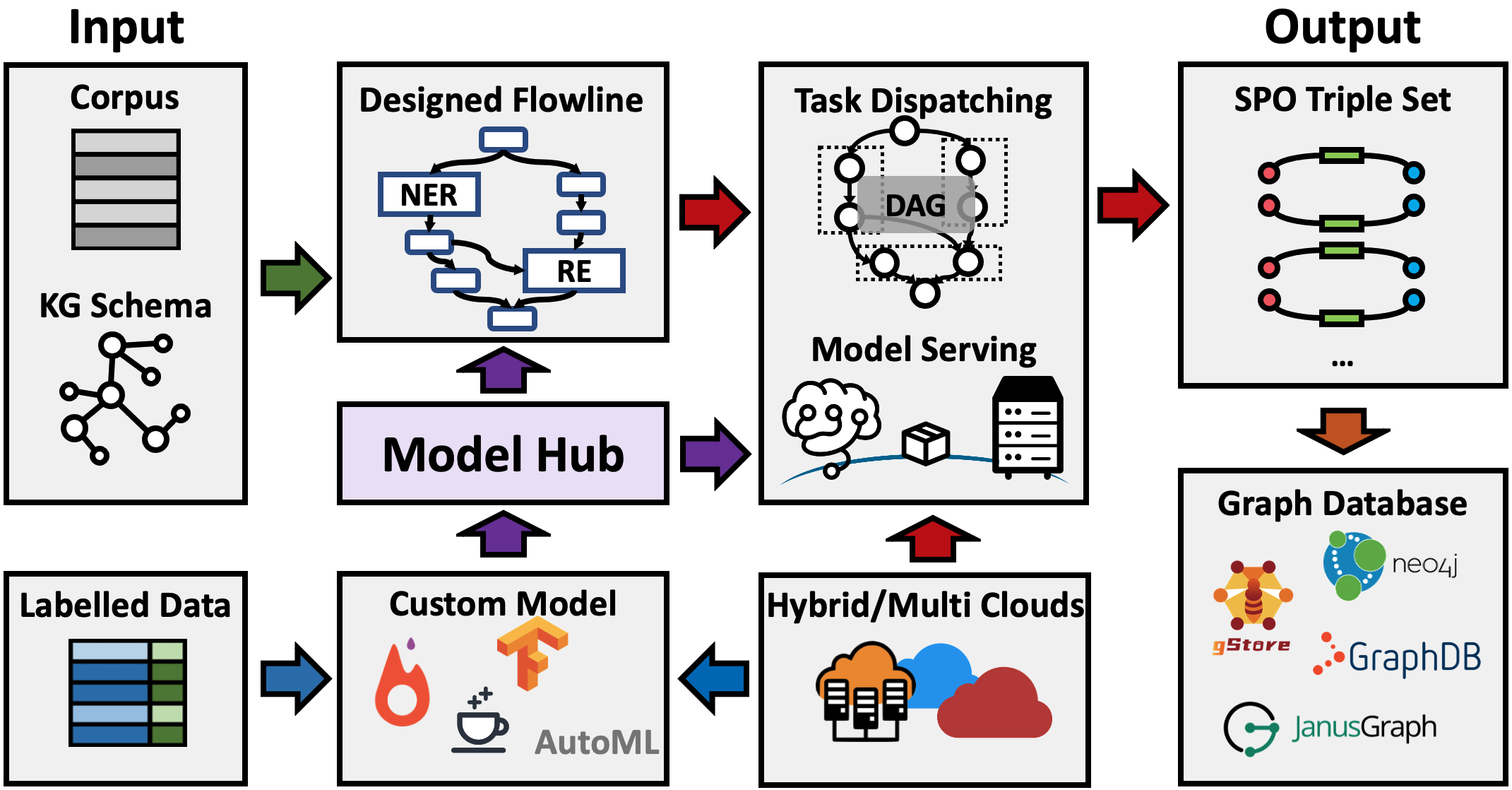}}
	\caption{The overview diagram of gBuilder.}
	\label{fig:overall}
\end{figure}

\Paragraph{System Design}. The overall architecture of gBuilder is shown in \Figure \ref{fig:overall}. gBuilder is modularity developed following the microservices architecture and deployed in multi-cloud environments for supporting multiple projects and users. 
gBuilder has several modules to implement basic functions like user, project, model and data management, and a cloud-based worker is dynamically launched for executing various KGC tasks.
For details on the system design, please refer to Appendix \ref{app:system} in our supplementary file.

\section{Unifed KGC Workflow: Flowline}\label{sec:flowline}




\subsection{Normalized IE Models}\label{sec:normalizedIE}

As Section \ref{sec:pre} described, there have been various studies on IE in recent years, including NER, RE, etc. However, the settings of different IE tasks are widely divergent. How to organize those algorithms and models with different settings, model structures, and output formats is a challenge. To normalize these IE models, we categorize them into two types according to the orientation of the IE models:

\noindent (1) \textit{Classification-based model} $CC$, which outputs a category label:
\begin{equation}\notag
CC\left(x_i\right) = \widehat{y}_{i},
\end{equation}
where $C$ represents classifier, $x_i$ denotes the $i$-th sample and $\widehat{y}_{i}$ denotes the $i$-th category label. 
The tasks or subtasks related to predicting the predicate are divided in this $CC$ category, including RE, attribute name extraction in AE, relation extraction in JE, etc. 
Even if the model originally outputs the spans, we will unify it into the $CC$ paradigm by building a label set dictionary. 

\noindent (2) \textit{Chuck Extraction-based model} $CE$, which outputs a set of fact phrase chunks:
\begin{equation}\notag
\setlength\abovedisplayskip{5pt}
\setlength\belowdisplayskip{5pt}
CE\left(x_i\right) = \widehat{Y}_{i} = \{ \widehat{y}_{i1}, ...,  \widehat{y}_{i|\widehat{Y}_{i}|}\},
\end{equation}
where $CE$ represents chunk extractor, and $\widehat{y}_{i}$ is the extracted fact from the sample $x_i$. The tasks or subtasks related to subject or object extraction are classified in this $CE$ category, including NER, attribute value extraction in AE, entity extraction in JE, etc. The above abstraction smooths out the logical differences caused by diverse models, such as sequence tagging, slot filling, and heuristic pattern matching.

\Paragraph{Built-in Models.} Following the above-mentioned normalized form principle, we prepared some built-in IE models for a startup.
In gBuilder, we use $(m, d_t, hp)$ to identify a deep learning-based model, where $m$ is the whole model, including the model structure and initialized parameters, $d_t$ denotes the training dataset, and $hp$ is the hyperparameter used to configure the training details. Implicitly, $d_t$ also defines the label set of a model. $d_t$ and $hp$ directly impact the tuned parameters of the model after training, and the performance of well-tuned model in the inference process. For the sake of ease of use, we train different models on different datasets in multi-linguals and multi-tasks. These models include classical and state-of-the-art models and popular repositories from academic or industrial open-source projects.
In order to make the maximum utilization of the precious supervised training datasets, and to provide users with as many diversified well-trained models as possible to reinforce model ensemble~\cite{zhou2021ensemble}, the ``matrix'' of built-in models approximately presents a combination of $M \times D_t (M=\bigcup m, D_t=\bigcup d_t)$, while the respective hyperparameters $hp$ are provided by the original author of models or fine-tuned by ourselves through automatic or manual tuning. With this strategy, we provide a large number of built-in models in the model hub of gBuilder and a part of employed datasets for well training built-in models, which are shown in Appendix \ref{app:built-in-model} in our supplementary file.

\Paragraph{Custom Model Training}
is a requisite for the adaption of domain-specific KGC, especially when the built-in models cannot meet the requirements of the ontology of the target domain. 
gBuilder provides abundant methods for custom model training. Users can invoke the custom model training function within gBuilder, or self-service registering a user-made model that is technology- and structure-agnostic to gBuilder, and providing inference service by the promissory interface for responding to the remote call from gBuilder system.

gBuilder provides three training modes for users: 
1) using a built-in well-trained model and uploading more data for continuous learning; 
2) using a built-in model structure, uploading training dataset and label sets, then training it from scratch. 
3) using pre-defined few-shot IE prompts for invoking LLMs' API to implement a prototype IE model.

Generally, the precision of KGC would be substantially improved after training a custom model using the corresponding supervised training dataset that conforms to the target KG domain.

\Paragraph{Custom Model Endpoint Register}.
gBuilder opens up an interface to advanced users, allowing users to register their self-serve models or algorithms and plug into the \texttt{flowline} in an agreed-upon format of data and endpoint state. It provides more flexibility to allow users to build and train specific models by themselves with better performance and manage complex data pipelines within \texttt{flowline}\footnote{We have released a repository (\url{https://github.com/pkumod/gbuilder-endpoint-example}) of a minimal implementation of custom model endpoint for guiding development.}.

\subsection{Unite Models and Ontology Merging}

In the practice of KGC, although training a custom model can provide better precision, it is often high-costly and even infeasible for domain engineers due to the lack of ML experience. A desirable approach is to unite build-in models to conform to the target KG ontology and use them directly without training custom models. However, well-tuned build-in models are often inconsistent with the target ontology, and the distribution of training data usually deviates from the target domain. To mitigate this phenomenon, we introduce and integrate multiple built-in IE models in gBuilder, and it is primarily geared towards choreography and fusion of these built-in models that are trained on different corpora with label sets.

\begin{figure}[t]
\centering{\includegraphics[width=.8\linewidth]{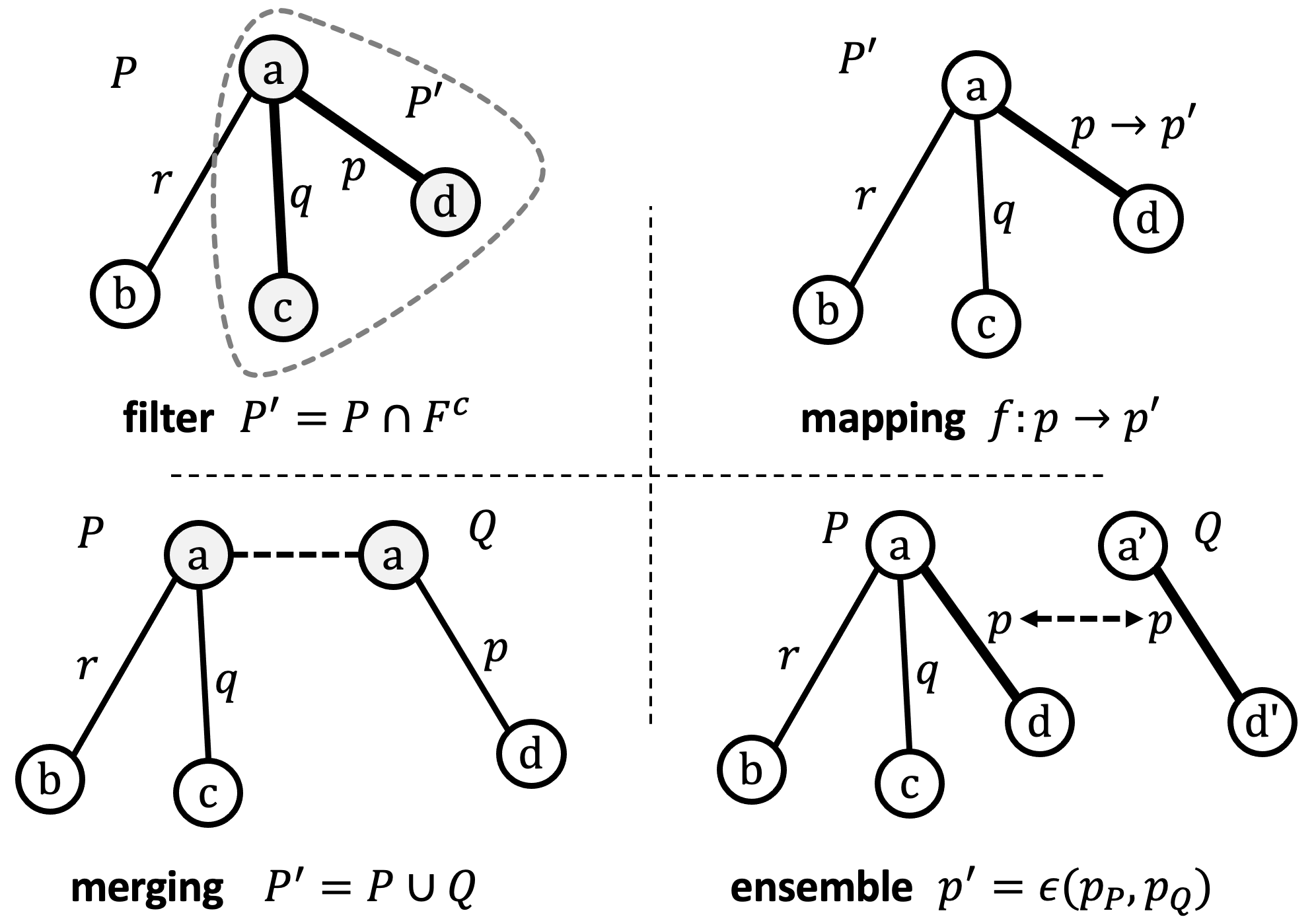}}
	\caption{The illustration of four primitive operations of ontology merging in gBuilder.}
	\label{fig:om}
	\vspace{-0.3cm}
\end{figure}

The output of deep learning IE models are usually discrete categories and can be treated as a flat structured ontology. The problem of uniting different IE models can be regarded as merging multiple ontologies into a target ontology. Traditionally, ontology merging needs to find the semantic relationship between concepts of different ontologies \cite{stuckenschmidt2002approximate, hitzler2005ontology, kalfoglou2002information}.
Within the scope of the flat ontology of IE models, regardless of the hierarchical and heterogeneous structure, ontology merging is relatively straightforward. 
We define four atomic operations to cover most ontology merging situations in \Figure \ref{fig:om}.
\textbf{filter:} pruning a set of classes or relations (defined as $F$ in the graph) that are not related to the target ontology.
\textbf{mapping:} literally mapping the concept name of the original ontology to the target ontology.
\textbf{merging:} directly merging multiple ontologies with non-overlapping concepts.
\textbf{ensemble:} adopt model ensemble approaches to address the overlapping concepts (including classes or relations) in multiple ontologies.

Through the flexible combination of these noncommutative operations to form a merging plan, gBuilder is sufficient for most situations of ontology merging. The merging plan is described by \texttt{flowline} (that will be discussed in Section \ref{sbusect:flowline}) and constructed by some of the built-in operators. After completing the merging plan at the ontology level, gBuilder will follow this plan to transform and merge knowledge instances output from IE models in the KGC process.

\Paragraph{Model Ensemble Tailored for IE Models.} 
In order to deal with the redundancy of multiple models outputting the analogical class during ontology merging and improve the prediction performance by integrating numerous well-tuned models with different architectures and trained on the diverse corpus, gBuilder introduces model ensemble for the inference process.

(1) \textit{Model ensemble applied in classification-based tasks.} Model ensemble is intrinsically suitable for classification-based models like relation classification. Therefore, the basic method of model ensemble can be used to solve the integration of those classification-based IE models. There are also two built-in ensemble operators in the gBuilder, one is based on vanilla majority vote:
\begin{equation}\notag
\widehat{y}_{i}=\operatorname{mode}\{C_{j}\left(x_i\right) | j \in \mathbb{N}_+, j < m \},
\end{equation}
where $m$ denotes the number of classifiers, $C_{j}\left(.\right)$ denotes the $j$-th classifier' prediction label, $\operatorname{mode}$ is for counting the most frequent labels.
The other is the score-based ensemble, which aims to gain the final predicted label by the weighted score sum among all classifiers for each label. The highest score label would be treated as the final results, and a threshold is introduced for filtering the unreliable predication. The score-based ensemble is shown as:
\begin{equation}\notag
\widehat{y}_{i}= \begin{cases}\underset{k}{\arg \max } \sum_{j=1}^{m} w_{j} p_{i, k}^{j}, & \text{if } \forall p_{i,k}^{j}>\delta \\ \text { reject } & \text {otherwise,}\end{cases}
\end{equation}
where $w_j$ denotes the weight of the $j$-th classifier (this weight is specified by the user, by default it is $1/m$). $p_{i, k}^{j}$ denotes the score that the $j$-th classifier predicting the $i$-th sample into the $k$-th class, $\delta$ is the threshold to accept the predicted label (by default it is $0.5$), if all prediction score below this threshold, the corresponding sample will be ignored, e.g., in relation classification task it will be marked as \textit{no relation} between entities.

(2) \textit{Model ensemble applied in extraction-based tasks.} The sequence tagging model is more diverse and complicated, and it is difficult to apply a unified score to measure the confidence coefficient of sequence tagging's prediction. 
%
%
Recently, some researches \cite{li2020survey} proposed sequence tagging-based model ensemble approaches based on encoder \cite{paccanaro2001learning} or tag decoder \cite{celikyilmaz2015investigation}. However, these inner-model ensemble methods require specific model structures and are not easy to transfer across different domains, and it is over fine grain and incompatible with gBuilder, which is based on the well-tuned model pipeline.
Therefore, based on our introduced \textit{CE} extraction paradigm, we implement ensemble operators outside of models by merging and deduplicating extracted chunks directly and automatically.
In this way, we can introduce multiple models in extraction-based tasks like NER, which improve the recall score as possible to minimize cascading errors for the overall knowledge extraction pipeline.

Inevitably, this manner would produce some exceptional cases such as redundant nested named entities. Those special cases are delivered to the post-processing, including knowledge graph quality control, entity disambiguation. Experimental results have shown that this manner does not harm the overall accuracy of knowledge instance extraction.

\Paragraph{Built-in Operators.}
For the implementation of uniting different models, besides \emph{filter}, \emph{mapping}, \emph{merging} and \emph{ensemble}, we propose a series of \emph{built-in operators} are preset in gBuilder, allowing users to build the \texttt{flowline} (Section \ref{sbusect:flowline}) conveniently. Appendix \ref{app:builtin} in supplementary file demonstrates the functions, inputs, and outputs of the most of built-in operators in details. These built-in operators can be flexibly combined to implement the features required by KGC.

\subsection{Flowline: Choreography of Operators}\label{sbusect:flowline}

With the built-in models for extracting information and the built-in operators for uniting these models, the KGC pipeline can be described in an abstract form, called \texttt{Flowline} in gBuilder, a combination of IE workflow and data pipeline.

\begin{definition}[Flowline]\label{def:flowline} \emph{Flowline} is a Directed Acyclic Graph (DAG) that abstracts KGC workflow in which (1) each vertex represents a IE model or an build-in operator. Note that each node is also called a \emph{task}; (2) an edge defines the data flow between the two corresponding tasks; (3) there are a single \emph{entry vertex} and a single \emph{exit vertex} that correspond to the start and the end of a flowline, respectively. 
\end{definition}


Once the flowline is determined, the resource estimation of each node in the workflow will also be determined, and the system can schedule such a DAG to ensure the entire flowline to be completed as soon. We will propose a DAG-partition strategy to optimize optimizing flowline schedule in the cloud-computing environment in Section \ref{sec:schedule}. 

\begin{figure}[t]
\centering{\includegraphics[width=0.8\linewidth]{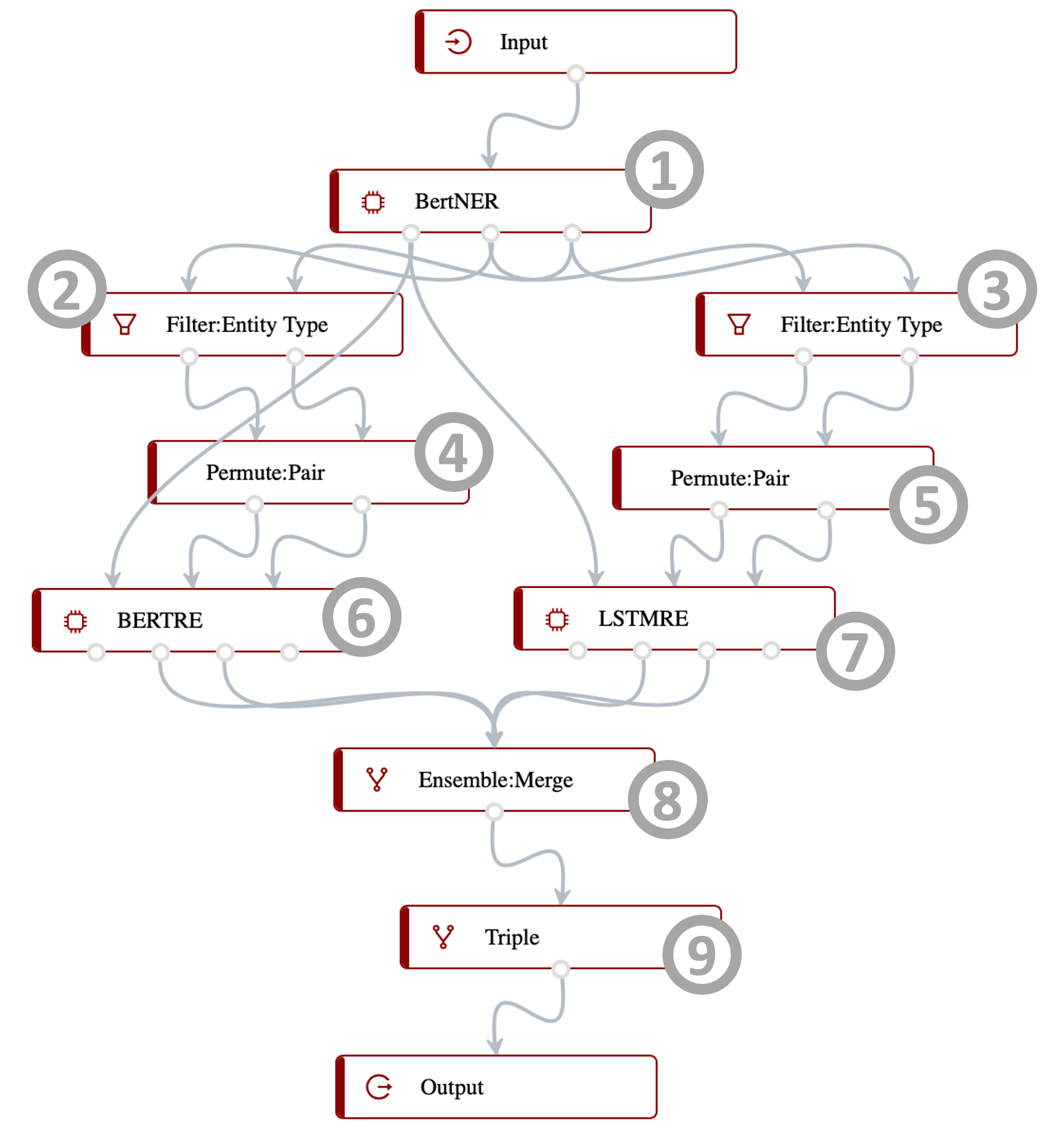}}
	\caption{Flowline example. For convenience, we have added numeric symbols to each task node.}
	\label{fig:flowline1}
	\vspace{-0.2cm}
\end{figure}

\Paragraph{Use Case of Designing a Flowline.}\label{sec:flowlinecase}
To explain the flowline’s design clearly, we conducted a case study with the example of extracting knowledge instances in \Figure~\ref{fig:kgc-exp}, and \Figure~\ref{fig:flowline1} shows the flowline we have designed for this purpose.

From a holistic perspective, to extract knowledge instances, we must first obtain the entities in the text. So we arrange the build-in BertNER model$_{(1)}$\footnote{The bracketed number represents the corresponding node in \Figure~\ref{fig:flowline1}.}, which is a high-performance NER model selected, with the ability to extract the person’s name, location, organization name, and some of the attributes (Date, Time, etc.), which meets the requirements of the pre-defined ontology.

After further research, we search in the model hub and found that an RE model named BERTRE$_{(6)}$ that specializes in identifying relations between company and person, as well as an RE model named LSTMRE$_{(7)}$ that can identify attributes relation of company founded date. 
Therefore, we chose to make branches off in flowline for identifying person-organization relations and organization-(founded date) attribute relations to exploit the specialty of these two models, respectively.

For this purpose, we employ filter nodes$_{(2,3)}$ to split the data by entity type and then transform the entities into entity pairs through “permute:pair” nodes before feeding entity pairs into the RE models$_{(6,7)}$.
After identifying relations and attributes, we use an Ensemble node$_{(8)}$ to integrate the results and then go through the triple constructor$_{(9)}$ to get the final structured knowledge instance in the form of RDF triples.

Note that this flowline represents a minimum viable example to demonstrate the primary KGC process. In the production environment, KGC pipelines usually employ substantially more complex flowlines, often involving over 30 nodes.

\section{Task Dispatching}\label{sec:schedule}

Handling a KGC project with multiple deep learning models requires large-scale heterogeneous computing resources (including CPUs and GPUs, etc.), and furthermore, providing such massive KGC services to numerous users poses a huge challenge to the computing resources. 
Therefore, gBuilder adopts cloud computing architecture and employs the auto-scaling feature to provide users with high-quality services. As a cloud distributed system, the scheduling of tasks will directly determine the scalability, Quality of Service (QoS), resource utilization, and the monetary cost of cloud procurement. This section focuses on task dispatching in gBuilder. 

\subsection{Micro-batch Distributed Cloud Computing}\label{sec:micro-batch}

To minimize the impact of the data transmission delay between cloud VMs (virtual machines) on the time overhead of the overall flowline, a micro-batch-based data layout is designed to reduce the proportion of transmission time within the overall flowline and a task-oriented column-based transmission is employed to reduce the data size at intermediate computing nodes.

\Paragraph{Data Layout.}
\begin{figure}[h]
\centering{\includegraphics[width=0.35\linewidth]{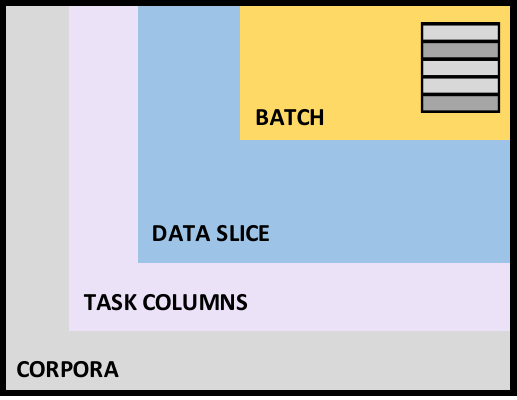}}
	\caption{The hierarchical data layout of gBuilder.}
	\label{fig:data-hierarchy}
	\vspace{-0.2cm}
\end{figure}
The data layout in gBuilder is defined as four levels, as shown in the \Figure \ref{fig:data-hierarchy}. 
\begin{itemize}[leftmargin=*,align=left]
\item \emph{Corpora} contains all documents of an entire project.
\item\emph{Task Column} is a set of data columns related to a single task, which only contains the required fields. We split the data into columns and then group/transmit the necessities for successor tasks to reduce the intermediate transmission data amount (e.g., a single \texttt{entity\_type\_filter} task only requires the input of \texttt{entity\_type} and only the corresponding data column from precursor tasks would be transferred to this computing node).
\item \emph{Data slice} is an essential data transmission unit (i.e., \textit{micro-batch}\footnote{It is named as ``data slice'' to avoid counter-intuitiveness with the concept of ``batch'' in batch learning.}) between a task (i.e., a vertex) and its successor in the flowline. The size of the data slice is determined by the system environment. Within the data slice, data rows can be order-preserved. Data slices could be distributed to different tasks and asynchronously invoked to save data transfer time \cite{warren2015big, spark}.
\item \emph{Batch} is identical to the concept of batch learning. By batching the data, the dedicated computing hardware (like GPU, TPU) can maximize the acceleration of matrix calculations to increase the speed of deep learning model inference or training \cite{cudnn}.
\end{itemize}



\subsection{Problem Formulation}
\begin{table}[t]
\small
    \caption{Common used notations in Flowline scheduling.}
    \label{tab:notation}
    \begin{tabular}{p{0.2\columnwidth}p{0.7\columnwidth}}\toprule
        Notations & Definitions\\
        \midrule
        $G=(V, E)$ & A DAG with a vertex set $V=(v_1, v_2, ...v_i)$ and an edge set $E=(e_1, e_2, ..., e_j)$\\
        $t_i$ & Task $t_i$ in Flowline, which is interchangeable with vertex $v_i$ in DAG $G$ \\
        $|D|$ & The size of corpora $D$\\
        $|slice|$ & The size of micro-batch data unit (data slice) \\
        $T$ & Task list of Flowline, $T=\bigcup t_i, T\equiv V$ \\
        $V^m, V^o$ & Sub-list of GPU- and CPU-intensive tasks/vertices \\
        $vm$ & Cloud Virtual Machine \\
        $N(v)$ & Neighbour node set of vertex $v$ \\
        $N_{in/out}(v)$ & Precursor and successor node set of vertex $v$ \\
        $w(v_i)$ & Weight of vertex (equal to the execution time of task $v_i$) \\
        $w(\overrightarrow{v_iv_j} )$ & The time of unit data transferring from task $t_i$ to $t_j$ \\
        $FT(t_i)$ & The finish time of task $t_i$ \\
        $M(G)$ & \emph{makespan}, the running time of processing a unit data for entire DAG $G$ \\
        \bottomrule
    \end{tabular}
    \vspace{-0.4cm}
\end{table}

\subsubsection{Flowline Model}
To model the whole processing time of a flowline (i.e., a DAG, see Definition \ref{def:flowline}), we introduce vertex weight $w(v_i)$ and edge weight $w(\overrightarrow{v_iv_j} )$ to represents the excepted computation time of task $t_i$ (corresponding to vertex $v_i$) and the micro-batch data transmission time from task $t_i$ to $t_j$. Note that we ignore edge weight between $v_i$ and $v_j$ (i.e., $w(\overrightarrow{v_iv_j} )=0$) if the two tasks are executed at the same VM. We only consider the data transmission time among different VMs.

gBuilder online applies for requested cloud resources to execute a flowline job. Given a flowline DAG $G(V,E)$, we assume that there are $|V^m|$ IE model vertices and $|V^o|$ operator vertices in $G$ and each IE model vertex needs a GPU card and each operator vertex needs a CPU core. To better motivate our task scheduling, we first consider the following ideal case.

\Paragraph{Ideal Case.} Given a flowline DAG $G$ over corpora $D$, in the ideal case, we assume that we can apply for a powerful virtual machine (VM) with enough GPU cards ($\geq |V^m|$) and CPU cores ($\geq |V^o|$) and we do not care the monetary cost. As mentioned earlier, we adopt the \emph{micro-batch} scheme (Section \ref{sec:micro-batch}) and one data slice is an data transmission unit. Therefore, the total processing time can be evaluated as follows:
\begin{equation}\label{def:timeIdeal}
Time(G,D) = \frac{{|D|}}{{|slice|}} \cdot M(G)
\end{equation}
where $|D|$ and $|slice|$ denote the data volume of the whole corpora $D$ and one data slice size, 
and $M(G)$ denotes \emph{makespan}, which refers to the running time of processing a data slice through the entire DAG $G$. $M(G)$ is equal to the \emph{finish time} (FT) of the exit vertex $t_\text{exit}$, where the \emph{finish time} for any vertex $v_i$ in DAG $D$ as follows:
\begin{equation}\notag
\small
FT(v_i)=
\begin{cases}
w(v_i), \hspace {2.5em} \text{Only if } v_i \text{ is the entry vertex in DAG $G$.} \\
w(v_i) + \max_{v_j \in N_{in}(v_i)}(FT(v_j) + w(\overrightarrow{v_i v_j})).
\end{cases}
\end{equation}
where $N_{in}(v)=\{u \in V, e_{(u, v)} \in E \}$ denotes the precursor vertices of $v$. Essentially, \emph{makespan} is weight sum of all vertices and edges along the \emph{critical path}\footnote{The longest weighted path in the DAG is critical path, which determines the total runtime of the workflow.} in $G$. Note that all tasks are resident at one VM in the ideal case, thus, all edge weights are zero in $G$.  

\Paragraph{Partition-based DAG Scheduling.} Obviously, the above assumption is not practical. First, there may not be a powerful VM  equipped with enough resources, such as GPU cards. Second, the price of one powerful VM with full-size capability is much higher than the total monetary cost of multiple VMs with low resources. We will discuss the pricing model in cloud service vendors later (Section \ref{sec:pricingmodel}). Therefore, a desirable solution is to employ the DAG partition strategy to balance the performance and the total monetary cost. 

\begin{definition}[Partition] A \emph{partitioning} of a given DAG flowline $G=(V,E)$ is an edge cut-based partition $\pi=\{\mathcal{G}^\prime_1,\mathcal{G}^\prime_2,...,\mathcal{G}^\prime_k\}$, where each $\mathcal{G}^\prime_i(\mathcal{V}_i,\mathcal{E}_i)$ is a small DAG and  $\mathcal{V}_i \neq \emptyset$, $\bigcup^k_{i=1} \mathcal{V}_i = V$ and $\mathcal{G}_i \bigcap \mathcal{G}_j = \emptyset, 1\leq i\neq j \leq k$.
\end{definition}

Given a partition $\pi=\{\mathcal{G}^\prime_1,\mathcal{G}^\prime_2,...,\mathcal{G}^\prime_k\}$ over a DAG flowline $G$, we apply for $k$ VMs and each one accommodates one sub-DAG $\mathcal{G}^\prime_i$. Due to the data transmission between different VMs, we should consider edge weights if the corresponding tasks are resident at different VMs. Formally, we have the following definition.

\begin{definition}[Partition-based Computation Graph] Given a partition $\pi=\{\mathcal{G}^\prime_1,\mathcal{G}^\prime_2,...,\mathcal{G}^\prime_k\}$ over a DAG flowline $G(V,E)$, the corresponding \emph{computation graph} $G^\prime (V^\prime, E^\prime)$ is defined as follows:
\begin{itemize}[leftmargin=*,align=left]
\item $G^\prime$ has the same vertex and edge sets with $G$;
\item Each vertex $v^\prime$ in $G^\prime$ has the same vertex weight as the counterpart in $G$;
\item Each edge weight $w(\overrightarrow{v^\prime_iv^\prime_j} )$ is defined as follows:
\begin{equation}
\small
\label{equ:comm_time}
w(\overrightarrow{v^\prime_iv^\prime_j})=
\begin{cases}
\sim 0, & \text{if $t_i$ and $t_j$ are in the same VM.} \\
L_{v^\prime_i, v^\prime_j}+\frac{\texttt{Data}_{ij}}{\beta_{i,j}} & \text{if $t_i$ and $t_j$ are in different VMs.}
\end{cases}
\end{equation}
where $\texttt{Data}_{ij}$ denotes the size of data slice transferred from $v_i$ to $v_j$, $\beta_{i,j}$ represents the bandwidth between two VMs and $L_{v^\prime_i, v^\prime_j}$ is the network latency; 
\end{itemize}
\end{definition}

Based on the partition-based computation graph $G^\prime$, we can know the whole processing time:
\begin{equation}\label{def:timepartition}
Time(G^\prime,D) = \frac{{|D|}}{{|slice|}} \cdot M(G^\prime)
\end{equation}
where $M(G^\prime)$ denotes the \emph{makespan} of partition-based computation graph $G^\prime$. 
Note that different partitioning over $G$ may lead to different $M(G^\prime)$. Therefore, we formally define our DAG-based partition problem. 

\begin{definition}[DAG Partition]\label{def:dagpartition}
The DAG partitioning problem is to find the \emph{optimal partition solution}  $\pi=\{\mathcal{G}^\prime_1,\mathcal{G}^\prime_2,...,\mathcal{G}^\prime_k\}$ for DAG $G$, which should satisfies the qualification $\varpi$ and minimize the objective function $J$, where 
\begin{enumerate}[leftmargin=*,align=left]
\item The qualification condition $\varpi$ refers to the resource constraint, i.e., each VM $M_i$ has no less than $|V_i^m|$ available GPU cards and $|V_i^o|$ CPU cores, $i=1,...,k$, where $|V_i^m|$ and $|V_i^o|$ denote the number of IE models and operators in sub-DAG $\mathcal{G}^\prime_1$ that is resident at VM $M_i$;
\item The optimization objective is defined as:
\begin{equation}\notag
    \arg \min_{\pi} J (G, \pi).
\end{equation}
\end{enumerate}
\end{definition}

In the context of cloud computing, the optimization function $J (G, \pi)$ should consider the total processing time (Definition \ref{def:timepartition}) as well as the monetary cost. We will discuss the pricing model as follows. 

\subsubsection{Cloud Computing Pricing Model.}\label{sec:pricingmodel}

The pricing model determines the monetary cost of gBuilder as a cloud-based system, which would significantly impact task scheduling designing \cite{portella2019statistical}. \emph{G4 instance}\footnote{\url{https://aws.amazon.com/ec2/instance-types/g4/}} in Amazon Elastic Compute Cloud (AWS EC2) is the industry’s most cost-effective GPU instance for machine learning inference and graphics-intensive applications among EC2 product family \cite{khan2022exploration}, therefore we take g4dn (Nvidia GPU series) as an example to investigate the quotation of GPU cloud instances in the pay-as-you-go (on demand) pricing manner.
We take the number of CPU cores and GPU cards as parameters and hypothesize that they are combined in a linear relation \cite{bukh1992art}, the price can be expressed as:
\begin{equation}\label{equ:price}
\small
    Price_{vm}(\#CPU,\#GPU) = \theta_1 \times \emph{\#CPU} + \theta_2 \times \emph{\#GPU}
\end{equation}
where $\#CPU$ and $\#GPU$ denote the number of CPU cores and GPU cards, $\theta_1$ and $\theta_2$ are two fitting coefficients, which can be calculated easily by statisticizing the quotation list of cloud vendor and solving the linear system.
\begin{table}[t]
\addtolength{\tabcolsep}{1pt}
  \centering\small
    \caption{Multiple linear regression equation of the CPU-GPU number applied to estimate the price of Amazon EC2 g4dn instance family, with $\theta_1=0.0565$ and $\theta_2=0.3$.}
    \label{tab:pricing}
    \resizebox{\columnwidth}{!}{
    \begin{tabular}{lccccl} \toprule
    \multirow{2}{*}{Instance} & GPU & CPU & Unit Price & Estimated & Error \\ 
    &num&cores&(per hour)&(per hour)&(\%) \\
    \midrule
    {g4dn.xlarge} & {1} & {4} & {\$0.526} & \$0.526 & 0\% \\
    {g4dn.2xlarge} & {1} & {8} & {\$0.752} & \$0.752 & 0\% \\
    {g4dn.4xlarge} & {1} & {16} & {\$1.204} & \$1.204 & 0\% \\
    {g4dn.8xlarge} & {1} & {32} & {\$2.176} & \$2.108 & 3.13\% \\
    {g4dn.16xlarge} & {1} & {64} & {\$4.352} & \$3.916 & 10.02\% \\
    {g4dn.12xlarge} & {4} & {48} & {\$3.912} & \$3.912 & 0\% \\
    {g4dn.metal} & {8} & {96} & {\$7.824} & \$7.824 & 0\% \\ \bottomrule
    \end{tabular}
    } 
    \vspace{-0.2cm}
\end{table}

Easily-accessible statistical and experimental results of instance unit price estimation are shown in Table \ref{tab:pricing}. As can be seen, \Equation \ref{equ:price} has a minimal error rate ($<$5\%) for most of the instances under the setting of $\theta_1=0.0565$ and $\theta_2=0.3$, where $\theta_1$ and $\theta_2$ can be regarded as the unit price of CPU and GPU in the AWS quotation, respectively, which also proves that CPU and GPU are the main factors affecting pricing in the cloud computing scenario\footnote{We also investigated more pricing schemes from cloud vendors such as azure, Tencent, aliyun, etc., which are all in accord with the same strategy.}. In a nutshell, the parameter estimated multiple linear regression and \Equation \ref{equ:price} can be used as the foundational element for controlling monetary costs when designing optimization objective.

\subsubsection{Scheduling Objective}
Let us recall our DAG partition problem (Definition \ref{def:dagpartition}) and we focus on the objective function $J$ as follows. Given a partition $\pi=\{\mathcal{G}^\prime_1,\mathcal{G}^\prime_2,...,\mathcal{G}^\prime_k\}$ over Flowline $G$, the corresponding partition-based computation graph is $G^\prime$. Obviously, the objective function $J$ should consider both the computational cost $\text{cost}_\text{com}$ (i.e., the total processing time) and the monetary cost $\text{cost}_\text{mon}$ (i.e., cloud resource procurement). 
The $\text{cost}_\text{com}$ is defined as the total processing time:
\begin{equation}\notag
    \text{cost}_\text{com} =Time(G^\prime,D) = \frac{{|D|}}{{|slice|}} \cdot M(G^\prime),
\end{equation}
and the total monetary cost for processing entire corpora can be denoted as:
\begin{align}\notag
     \text{cost}_\text{mon} &= \sum_{i=1}^{i=k}{Price({vm_i})} \times Time(G^\prime, D) \\\notag &= \frac{{|D|}}{{|slice|}} \cdot \sum_{i=1}^{i=k}{Price({vm_i})} \cdot M(G^\prime)
\end{align}
where $Price(vm_i)$ is the purchasing cost of VM $vm_i$ and $Time(G^\prime, D)$ is the billing duration, which equals the total processing time of corpora\footnote{Delay of cloud instance acquisition and termination is ignored because this delay is negligible compared with the total KGC process.}.

%
Obviously, $\frac{{|D|}}{{|slice|}}$ is fixed when finding the optimal partition. Thus, we can only consider the unit price for processing a data slice when defining the objective function. Furthermore, simultaneous optimizing both $\text{cost}_\text{com}$ and $\text{cost}_\text{mon}$ is a typical multiobjective optimization (MO) problem, and it is hard to find a Pareto-optimal solution. Motivated by multi-criteria workflow scheduling problem \cite{su2013cost, kurowski2006grid}, we use weighted aggregation to formulate the MO into a single objective problem by combining the computational cost and monetary cost:
\begin{align}\label{equ:objective}
    \text{minimize} \indent &J (G, \pi) = \eta \text{cost}_\text{com} + (1 - \eta) \text{cost}_\text{mon}\\ \nonumber
   \text{subject to} \indent &\pi \in \varpi, 0 \leq \eta < 1,
\end{align}
where $\eta$ is the preference factor that indicates the relative importance of the decision-making, and it depends on the decision maker's trade-off between QoS and the monetary cost. Unfortunately, this optimization problem is NP-hard as given in the following theorem.  


\begin{theorem}\label{thm}
Given a DAG $G$ for a valid Flowline, finding the optimal graph partition to minimize \Equation \ref{equ:objective} has no polynomial-time approximation algorithm with a finite approximation factor unless $P=NP$\footnote{This theorem can be easily proofed by reducing to Minimum Makespan Scheduling problem, and the detailed proof is given in the Appendix \ref{app:hard}.}.
\end{theorem}

\subsection{Proposed Solution}
Due to the hardness of finding the optimal DAG partition, considering IE computation features in KGC problem, we propose a heuristic algorithm for scheduling Flowline in a distributed cloud computing environment.

\subsubsection{Algorithm Idea}\label{sec:prior}
Note that we do not intend to conquer the DAG partition problem in a general case in this paper. We first study IE computation features in KGC problem and we propose our solution based on the prior knowledge.
Considering the pricing model, we can use \Equation \ref{equ:price} to estimate the monetary cost of cloud procurement and resource utilization for gBuilder's distributed computing workers.
In most cases, the range of CPU-GPU ratio required by the gBuilder worker is typically from $2$ to $6$ (e.g., an IE model with 1 filter, 1 mapper, 1 ensemble operator, 1 data integration operator, 1 constructor operator).
It supports us to choose an instance type with a CPU-GPU ratio as close to the requirements of gBuilder as possible. Cloud vendors tend to bundle rich additional resources in order to drive users to choose a high-end instance with multiple GPU cards, which will result in low resource utilization in our application scenarios. 
For example, if we purchase VMs in Table \ref{tab:pricing} for gBuilder, the g4dn.metal instance would cause 34.66\% to 57.77\% of monetary waste. Therefore, with the vast resource utilization deviation, we prefer to purchase the basic instance type with a single GPU and an appropriate number of CPU cores like g4dn.xlarge and g4dn.2xlarge to avoid superfluous CPU resources cost.
Other prerequisites that should be considered when developing the scheduler include:
\begin{enumerate}[leftmargin=*,align=left]
	\item Workloads that serve deep learning-based IE models consume most of the time slot in the pipeline.
	\item The latency of network communication between cloud instances is non-negligible and heavily affects micro-batch processing execution efficiency. Conversely, the inner-instance communication can be negligible.
	\item Empirically, An GPU-heavily deep learning IE model usually succeeds a series of lightweight operators (such as filters, mappers, etc.) for postprocessing. These operators are mainly consuming CPU resources.
	\item A GPU cloud server always provides superfluous computing capacity such as redundant CPU cores, etc.
	\item A deep learning-based IE model could be regarded as exclusively occupying all of the single GPU resources but only occupying 1 CPU core of the capacity for data loader.
	\item Resource demand of ``lightweight'' operators can be regarded as 1 CPU core.
\end{enumerate}

\subsubsection{Heuristic Algorithm} Building on the above ideas, we can propose a heuristic algorithm to optimize the objective in \Equation \ref{equ:objective} and solve the hardness problem of flowline scheduling.
To this end, we design a two-step algorithm: 
first, ensure the compounded tasks should not be separated as the unpartitionable clusters; then, determine the partition strategy and the number of clusters based on a greedy approach.

For convenience, we denote GPU-intensive tasks like IE models as $V^m$ and CPU-only tasks like lightweight operators as $V^o$. The unpartitionable ``compound'' is denoted as $P_i$.

\Paragraph{Step 1. The Compounding Phase.} Motivated by intuitive assumption 1, we construct a compounding algorithm based on bundling the GPU-intensive task and its subsequence CPU-only tasks. These subpopulations are self-contained and \emph{should} be clustered; delimiting boundaries within those \emph{compounded tasks} would only bring extra communication cost without any profit.
We summarize the GPU-task (i.e., the IE models) guided compounding in Algorithm \ref{alg:compound}. 

\begin{algorithm}[ht]
\small
\caption{GPU Task-Guided Compounding}\label{alg:compound}
\SetKwInOut{Input}{Input}
\SetKwInOut{Output}{Output}
\SetKwRepeat{Repeat}{repeat do}{until}
  \Input{A Flowline $G=(V,E)$}
  \Output{A Set of Compounded Subpopulations $P = \{P_1, P_2, ..., P_{|V^m|}\}$}
  \BlankLine
  \emph{$P \leftarrow \emptyset$}\\
  \For{$t_i \in \{t|t \in V^m\}$}{
  \emph{$P_i = \{t_i \}$}\\
  \Repeat{visited all successors of nodes $t'$ in $P_i$ in $G$}{
  	\For{$t' \in N_{out}(P)$}{
  	\uIf{$t' \in V^o$ \textbf{and} $N_{in}(t') \subseteq P_i$}{
	   	\emph{$P_i.join(t')$}
  	} 
  	}
  }
  }
 \end{algorithm}
 
\Paragraph{Step 2. Cloud Resources Pre-allocation.}
To determine the procurement plan of cloud resources and allocate the tasks to the cloud instances, we dig into pricing strategy of the cloud vendor 
and set up a simple polynomial function between the minimum makespan of KGC process and the procurement pricing of cloud resources:
\begin{equation}\label{equ:cost2makespan}
\setlength\abovedisplayskip{5pt}
\setlength\belowdisplayskip{5pt}
    y = g(x) = a + b (x-c)^{-1}, (a,b,c,x > 0)
\end{equation}
where $a$,$b$, and $c$ are the coefficients to fit which are dependent on the pricing strategy of cloud vendors, $x$ denotes the planned unit price of a task, $y$ denotes the estimated minimum makespan of the task according to the planned expenditure.\footnote{Due to space limitations, the detail derivation is attached in the Appendix \ref{app:cost2makespan}}.

Substituting the hypothesis polynomial function relationship between planned unit price and minimum makespan into the objective function, it can be rewritten as:
\begin{equation}\notag
    J = \eta g(x) + (1-\eta) x \cdot g(x),
\end{equation}
it is a convex function with a global minimum at $x_0 = \sqrt{ \frac{b}{a} \frac{\eta}{1 - \eta} } + c$, which means the approximate optimal solution of $J$ can be obtained if a combination of cloud instances whose summation of unit price is close to $x_0$ is purchased. This is a classic unbounded knapsack problem that can be solved in various manners (e.g., dynamic programming. Furthermore, it is easy to find a brute-force solution because of the small scale of combinations in this scenario). 
Note that if there are multiple procurement plans for the same price, it is preferable to choose a combination of high-end instances than multiple divided low-end instances if possible, and the coefficients can be calculated by a few observations by \emph{warm up}\footnote{In warm-up period, tasks would be executed serially in topological order for processing a few data in a basic cloud instance, the makespan of tasks can be measured.}, and simple regression analysis methods like non-linear least squares can be employed.
For example, assuming the user expect a balance $\eta = 0.5$, and deploy a Flowline (as \Figure \ref{fig:flowline1}) in qCloud (with setting of $a=4.17$, $b=5.15$ and $c=23.96$), the expected price of optimal scheduling should close to $\sqrt{\frac{b}{a} \frac{\eta}{1-\eta}} + c = 25.08$, which means the purchasing plan with most close price should be applied.

With the purchased cloud instances \emph{VM}, it must satisfy the constraints of Definition \ref{def:dagpartition} and the resource requirement can be satisfied. For convenience, we denote the number of \emph{VM} as $k = |\emph{VM}|$, and the CPU and GPU capacity of the $i$-th VM are denoted as $CPU_i$ and $GPU_i$, respectively. The list of \emph{VM} is sorted by their GPU capacity in descending order.

\Paragraph{Step 3. Greedy Graph Partition.}
Motivated by Assumption 2, we construct a greedy graph partition strategy for minimizing the overall cost in \Equation \ref{equ:objective}. This phase is towards the compounded clusters and the omissive orphaned nodes that weren't assigned into any cluster in the previous steps.
Those orphaned nodes mainly include: 
(1) the node of an aggregation-type task demands multiple data flows produced by different GPU-intensive tasks.
(2) the tasks that must be isolated due to exhaustion of computing resources of the corresponding GPU cloud server.

For each vertex (including orphaned tasks and GPU task-guided compounded clusters) in the flowline, we use objective
\begin{equation}\notag
\setlength\abovedisplayskip{2pt}
\setlength\belowdisplayskip{2pt}
    \arg \max_i \{ |\mathcal{G}^\prime_i \bigcap N(v^\prime) | \}
\end{equation}
to maximize the number of neighbor nodes of vertex $v^\prime$ greedily, which can maximize the number of inner-partition edges and minimize the number of inter-partition edges as \Equation \ref{equ:comm_time}. The proposed greedy graph partitioning algorithm is summarized in Algorithm \ref{alg:partition}.
\begin{algorithm}[ht]
\small
\caption{Greedy Graph Partitioning}\label{alg:partition}
\SetKwInOut{Input}{Input}
\SetKwInOut{Output}{Output}
\SetKwRepeat{Repeat}{repeat do}{until}
  \Input{Flowline $G=(V,E)$, \\Compounded tasks set $P = \{P_1, P_2, ..., P_{|V^m|}\}$}
  \Output{Partition strategy $\pi = \{\mathcal{G}^\prime_1, \mathcal{G}^\prime_2, ..., \mathcal{G}^\prime_k\}$}
  \BlankLine
  \emph{$\mathcal{G}^\prime_{1\rightarrow k} \leftarrow \{\emptyset\}$} \tcp{initialize partitions}
  \For{$v^\prime \in P$}{
     \emph{$\text{indices} \leftarrow \arg \text{sort}_i \{ |\mathcal{G}^\prime_i \bigcap N(v^\prime) | \}$} \\
     \For{$n \textbf{ of } \text{indices}$}{
        \uIf{$ |\mathcal{G}_n^\prime \bigcap V^m | + | v^\prime \bigcap V^m | \leq GPU_i $}{
      	\emph{$\mathcal{G}^\prime_n$.join($v^\prime$)} \\
      	\textbf{break} \\
  	    }
     }
  }
  \For{$v^\prime \in \{v^\prime | v^\prime \in V^o$ \text{and} $v^\prime \notin P\}$}{
     \emph{$\text{indices} \leftarrow \arg \text{sort}_i \{ |\mathcal{G}^\prime_i \bigcap N(v^\prime) | \}$} \\
     \For{$n \textbf{ of } \text{indices}$}{
        \uIf{$ |\mathcal{G}_n^\prime \bigcap V^o | + | v^\prime \bigcap V^o | \leq CPU_i - GPU_i $}{
      	\emph{$\mathcal{G}^\prime_n$.join($v^\prime$)} \\
      	\textbf{break} \\
  	    }
     }
  }
 \end{algorithm}\vspace{-0.5cm}
 
With the determined $k$ size of partitions and the VMs, the computing subgraph $\mathcal{G}^\prime_i$ can be assigned to the corresponding VM$_i$.

\Paragraph{Corner Case.} Suppose a CPU-only flowline (e.g., the flowline is composed of operators without any deep learning model). It would skip the compounding phase. The scheduler would remain this flowline integral and purchase one VM with adequate CPUs.

\Paragraph{Time Complexity.} Algorithm \ref{alg:compound} can substantially reduce the search space for the subsequent graph partition phase. Its complexity mainly depends on the number of GPU-intensive task $|V^m|$ and the average out-degree $|\bar{N}_{out}(V^m)|$, denotes as $O(|V^m|^2 |\bar{N}_{out}(V^m)|)$.
In Algorithm \ref{alg:partition}, the \emph{argsort} procedure can be regarded as $O(k \log k)$, and the overall complexity of Algorithm \ref{alg:partition} is $O(|V| k^2 \log k)$.

\section{Experiment}

\subsection{End-to-end KGC Performance Evaluation}

\begin{table*}[!ht]
\addtolength{\tabcolsep}{2pt}
  \centering
  \caption{Detail results of triple extraction on various benchmark datasets. $\dag$ marks the results reported by the original paper.}
  \label{tab:ie-exp}%
  \resizebox{\linewidth}{!}{
    \begin{tabular}{rccccccccccccccc}
    \toprule
    \multicolumn{1}{c}{\multirow{2}[2]{*}{Models}} & \multicolumn{3}{c}{NYT} & & \multicolumn{3}{c}{WebNLG} & & \multicolumn{3}{c}{TACRED} & & \multicolumn{3}{c}{DUIE2.0} \\
\cmidrule{2-16}  & P     & R     & F   &  & P     & R     & F   &  & P     & R     & F   &  & P     & R     & F     \\
    \midrule
    CasRel & 89.7$^\dag$ & 89.5$^\dag$ & 89.6$^\dag$ && 93.4$^\dag$ & 90.1$^\dag$ & 91.8$^\dag$ && -     & -     & -     && 80.5  & 75.0    & 77.7  \\
    ChatGPT & 63.2 & 60.5 & 61.8 && 65.5 & 60.8 & 63.1 && 42.6 & 50.3 & 46.1 && 50.2 & 57.6 & 53.6  \\ \midrule 
    \emph{(M1)}    & 69.2 & 65.4 & 67.2 && 72.3 & 72.1 & 72.2 && 59.3  & 61.6  & 60.4  &&  68.1 & 66.8 & 67.4    \\
    \emph{(M2)}    & 69.9 & 68.6 & 69.2 && 74.0 & 73.8 & 73.9 && 62.3  & 61.5  & 61.9  &&  69.3 & 68.9 & 69.1    \\
    \emph{(M3)}    & 73.7 & 74.3 & 74.0 && 76.1 & 76.3 & 76.2 && 63.8  & 63.6  & 63.7  &&  69.9$^\ddag$ & 70.2$^\ddag$ & 70.0$^\ddag$    \\
    \emph{(M4)}    & 71.6 & 73.4 & 72.5 && 74.6 & 74.2 & 74.4 && 63.2  & 61.7  & 62.4  &&  69.5 & 69.1 & 69.3     \\
    \emph{(M5)}    & 60.4 & 49.2 & 54.2 && 65.2 & 61.6 & 63.3 && 58.9  & 57.5  & 58.2  &&  49.8 & 55.1 & 52.3    \\
    \emph{(M6)}    & -    & -    & -    && -    & -    & -    && -     & -     & -     &&  71.3 & 70.6 & 70.9     \\
    \emph{(M7)}    & -    & -    & -    && -    & -    & -    && -     & -     & -     &&  73.7 & 74.9 & 74.3    \\
    \emph{(M8)}    & 68.9 & 69.1 & 69.0 && 72.5 & 74.0 & 73.2 && 60.1  & 67.7  & 63.7  &&  66.5 & 67.3 & 66.9     \\
    \bottomrule
    \end{tabular}
    }\vspace{-0.2cm}
\end{table*}%

\subsubsection{Experimental Setup} To examine the feasibility of serving and ensembling multiple IE models within the gBuilder framework, we conduct experiments on various NLP tasks and benchmark datasets. These experiments evaluate the effectiveness of individual IE models as well as the overall performance of the KGC pipeline in gBuilder.

\Paragraph{Evaluation Criteria.} We use Precision (P), Recall (R), and F1 score (F) to evaluate the performance of overall triple extraction. 
In KGC scenario, a triple is regarded as ``correct'' only if its entities, relation, and attribute are all exactly matched to gold standard \cite{wei2020CasRel}.

\Paragraph{Benchmark Datasets.} We conduct experiments on four benchmark datasets in multi-language, which are all commonly used in IE research. We give a brief introduction for each dataset:
\begin{itemize}[leftmargin=*,align=left]
\item \emph{NYT}~\cite{sandhaus2008new} dataset is a public IE benchmark in English. It consists of about 1M samples from New York Times articles with 24 kinds of relations. For fair comparison, we follow the setting from \cite{zeng2018extracting}, with selected 56,195 samples for training, and 5,000 samples for validation and test, respectively.
\item \emph{WebNLG}~\cite{web_nlg} is a dataset for Natural Language Generation task and adapted for evaluating triple extraction tasks. It contains 5,019 samples for training, 500 and 703 samples for validation and testing, respectively.
\item \emph{TACRED}~\cite{zhang2017tacred} is a large-scale relation extraction dataset in English, constructed by text from newswire and web collections. It has 41 types of relations and contains 68,124 training examples and 15,509 test examples. 
\item \emph{DUIE 2.0}~\cite{li2019duie} is an industrial-scale Chinese IE benchmark dataset in the general field, containing more than 11,000 training data and abundant annotation. We flatten the multiple slots schema, and only the primary object slot is regarded as the golden standard.
\end{itemize}
\Paragraph{Models.} We design 10 flowlines including joint and pipeline extraction models and then conduct KGC experiments upon them to verify the KGC performance of our system. All the IE models in flowlines are finetuned in the target dataset for a fair comparison. The details of the flowline design are described below:

\noindent (1) \textit{Joint Models}: We employ recent-years SOTA JE model CasRel~\cite{wei2020CasRel} as strong baselines, which can demonstrate the ability of our system to organize the jointly IE models.

\noindent (2) \textit{Large Language Models}: We introduce the recent popular few-shot LLM extractor as a fashion baseline. Specifically, we use $10$ demonstration samples with prompting template like~\cite{agrawal-etal-2022-large} and access the ChatGPT API to get response.

\noindent (3) \textit{Pipeline Models}: The pipeline models mainly consist of two stages, NER and RE, and each stage can integrate multiple models. We use the NER($\cdot$) and RE($\cdot$) notation to identify the task of the corresponding model. Simple operators such as filters, mappers, and integrators in the flowline are omitted. The conducted pipeline models are as follows:
\begin{itemize}[leftmargin=*,align=left,topsep=2pt,parsep=0pt,partopsep=2pt]
\item \emph{(M1)}: NER(BERT$_{base}$) + RE(BERT$_{base}$)
\item \emph{(M2)}: NER(BERT$_{large}$) + RE(BERT$_{large}$)
\item \emph{(M3)}: NER(RoBERTa) + RE(RoBERTa)
\item \emph{(M4)}: NER(BERT$_{large}$) + RE(MTB)
\item \emph{(M5)}: NER(LSTM-CRF) + RE(BiGRU-Att)
\item \emph{(M6)}: NER(FLAT) + RE(BERT$_{wwm\_ext}$)
\item \emph{(M7)}: NER(FastNER$_{large}$ + FLAT) + RE(BERT$_{wwm\_ext}$)
\item \emph{(M8)}: NER(LSTM-CRF + BERT$_{base}$) + RE(BERT$_{base}$)
\end{itemize}
The combination of these baselines can be used to validate gBuilder's support for PLMs \emph{(M6, M7)}, multilingual \emph{(M6, M7)}, different model architectures \emph{(M4, M5)} and parameter-scale \emph{(M5)}, respectively.
%
The detail of models is listed in Appendix \ref{app:built-in-model} of supplementary file.
All the baseline models, LLMs and pipelines described above can be expressed and organized in gBuilder Flowline and compared fairly. By comparing the compositions of these models, the KGC performance of gBuilder can be comprehensively evaluated.

\subsubsection{Experimental Result}

The experimental results are shown in Table \ref{tab:ie-exp}. 
Overall, we can observe that all introduced models and flowlines organized by gBuilder (including CasRel and ChatGPT) get effective results on the 4 benchmark datasets, proving the capacity of gBuilder in organizing IE models for KGC, and validating the feasibility of our designed model normalization approach, model uniting and ontology merging methods.

Comparing pipeline models, combinations with PLM-based IE models (\emph{M1-M3}) outperform combinations of traditional models (\emph{M5}). More advanced PLMs (\emph{M3, M6, M7}) outperform base PLMs (\emph{M1, M8}). This aligns with related NLP research \cite{qiu2020pre}.
Language-specific models (\emph{M6, M7}) significantly outperform others on corresponding language datasets, showing linguistics and language characteristics improve IE models \cite{li-etal-2022-enhancing}.
\emph{(M7)}, adding FastNER$_{large}$, improves F1 by 3.4\% and recall by 4.3\% over \emph{(M6)}, showing ensemble models improve KGC pipelines. 
\emph{(M8)}, adding LSTM-CRF NER, improves average recall by 3.05\% over \emph{(M1)}, showing additional lightweight models improve large model recall.
The above experimental results on extraction performance evaluation demonstrate the capability of gBuilder in organizing IE models for KGC, and provide insight into the design of the KGC Flowline under different circumstances.

In summary, the evaluation on the benchmark datasets demonstrates the feasibility of gBuilder in organizing the IE model pipeline and implementing the KGC capability. Regardless of the model's parameter scale, model architecture, language, or even using third-party APIs to implement extractors, the design and execution of KGC processes can be implemented in gBuilder to make best use of the advantages and bypass the disadvantages of various models.

\subsection{KGC Pipeline Scheduling Evaluation}

\subsubsection{Experimental Setup}

To verify the performance of the proposed scheduler, we organized experiments on real-world flowlines. For evaluating the scheduling performance in the actual production environment, we collected 5 representative KGC Flowlines from beta test users and ourselves, which contains a different number of models and operators.
To facilitate the measurement, we use qCloud's GN10Xp series as the optional instances for the experiments. To obtain the effect of our scheduling algorithm under diversified scenarios, the $\eta$ is adjustable and we try to explore as many available optimal solutions as possible. We can observe the trade-off between makespan and monetary cost in such setting.
A subset of NYT datasets is adopted as the material to be processed by random sampling of 8,000 sentences, and 200 sentences for warming up the scheduler, and employed several built-in fine-tuned models for IE (the accuracy of the IE model is not needed to consider in this evaluation). For convenience, the acquisition and termination delay of cloud instances are ignored.

\Paragraph{Baseline Algorithms.}
We compare our scheduling algorithm with several baseline algorithms that target the multi-objective optimization problem~\cite{zhang2014multi, durillo2014multi, zhou2019minimizing}, including:
\begin{itemize}[leftmargin=*,align=left,topsep=1pt,parsep=0pt,partopsep=1pt]
\item \emph{MOHEFT}~\cite{Durillo2012MOHEFTAM} developed from the famous list-based scheduling approach HEFT (Heterogeneous Earliest Finish Time), extending HEFT to address the MO problem. It is employed to optimize makespan and economic cost in cloud computing~\cite{durillo2014multi}, and provide users with a set of optimal solutions for selecting the one that most satisfies users' requirements.
\item \emph{SPEA2}~\cite{Zitzler2002SPEA2IT} (Strength Pareto Evolutionary Algorithm) is an evolutionary algorithm that uses tournament selection to select the new population for the solution with the highest strength value. SPEA2 is often applied to explore the Pareto front for MO problems, e.g., to schedule workflow with trading-off execution time and execution price~\cite{yu2007multi}.
\item \emph{OMOPSO}~\cite{RC05} is a multi-objective Particle Swarm Optimization (PSO) algorithm to simulate the particle swarm flying through the hyper-dimensional search space to explore optimal solutions. It also can be utilized to plan the MO scheduling that considers the cost and job execution time~\cite{kaur2018novel}.
\end{itemize}
All these algorithms are implemented from open source projects~\cite{BENITEZHIDALGO2019100598}\footnote{https://github.com/leimiaomiao/Multi-Objective-Workflow-Scheduling}$^,$\footnote{https://github.com/ZhaoKe1024/IntelligentAlgorithmScheduler} and we adapt and modify those MO algorithms for scheduling. Multiple solutions generated by these baseline MO algorithms are also kept for comparison, and invalid scheduling plans (e.g., unable to meet resource requirements or the economic cost is too much higher than ours) are rejected before execution.
The parameters of the algorithms keep the same as the previous works~\cite{durillo2014multi, zhou2019minimizing}.

\subsubsection{Experimental Results}
\begin{figure*}[ht]
\centering
\subfloat[]{\includegraphics[width=2.35in]{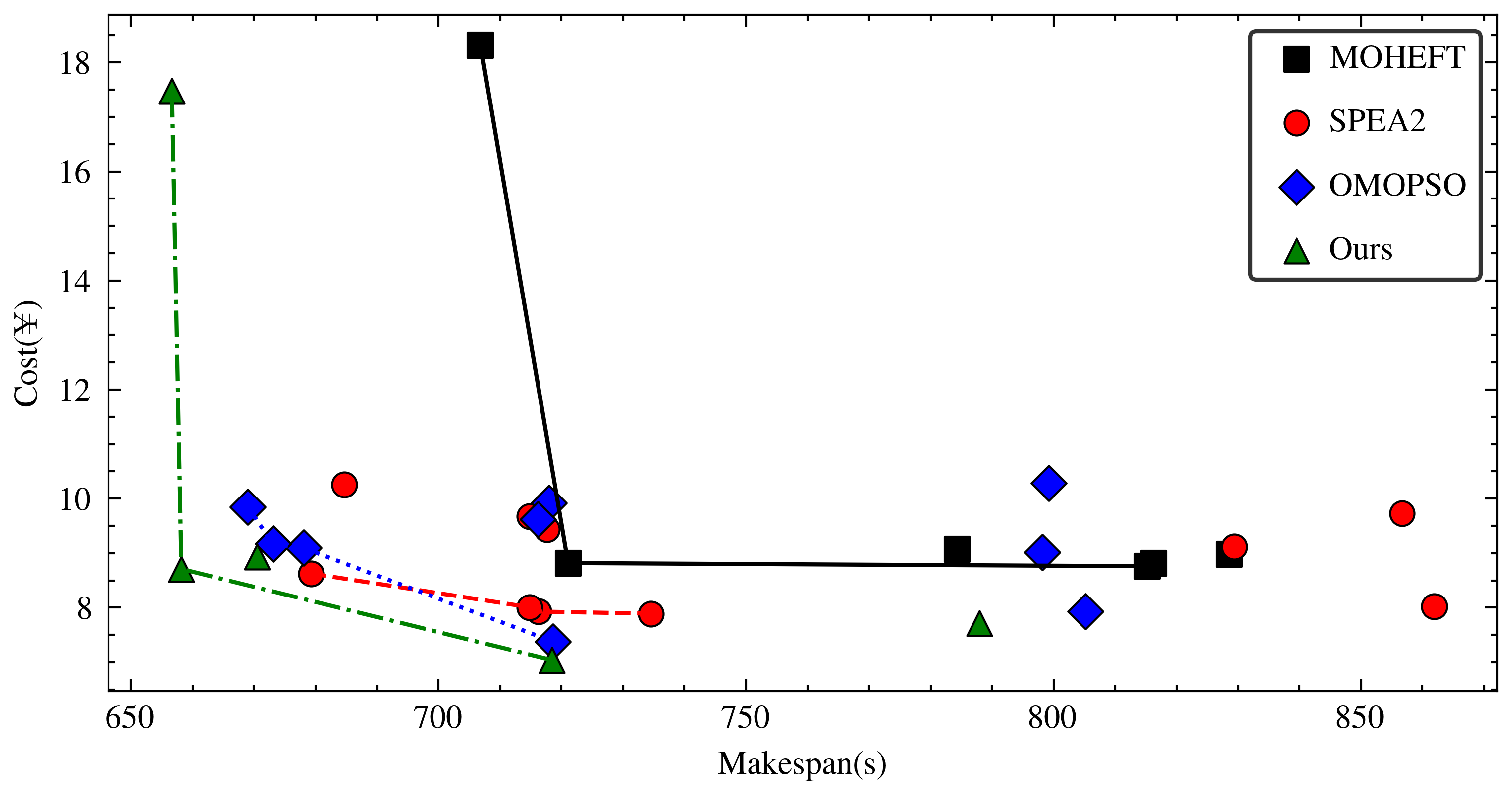}}
\subfloat[]{\includegraphics[width=2.35in]{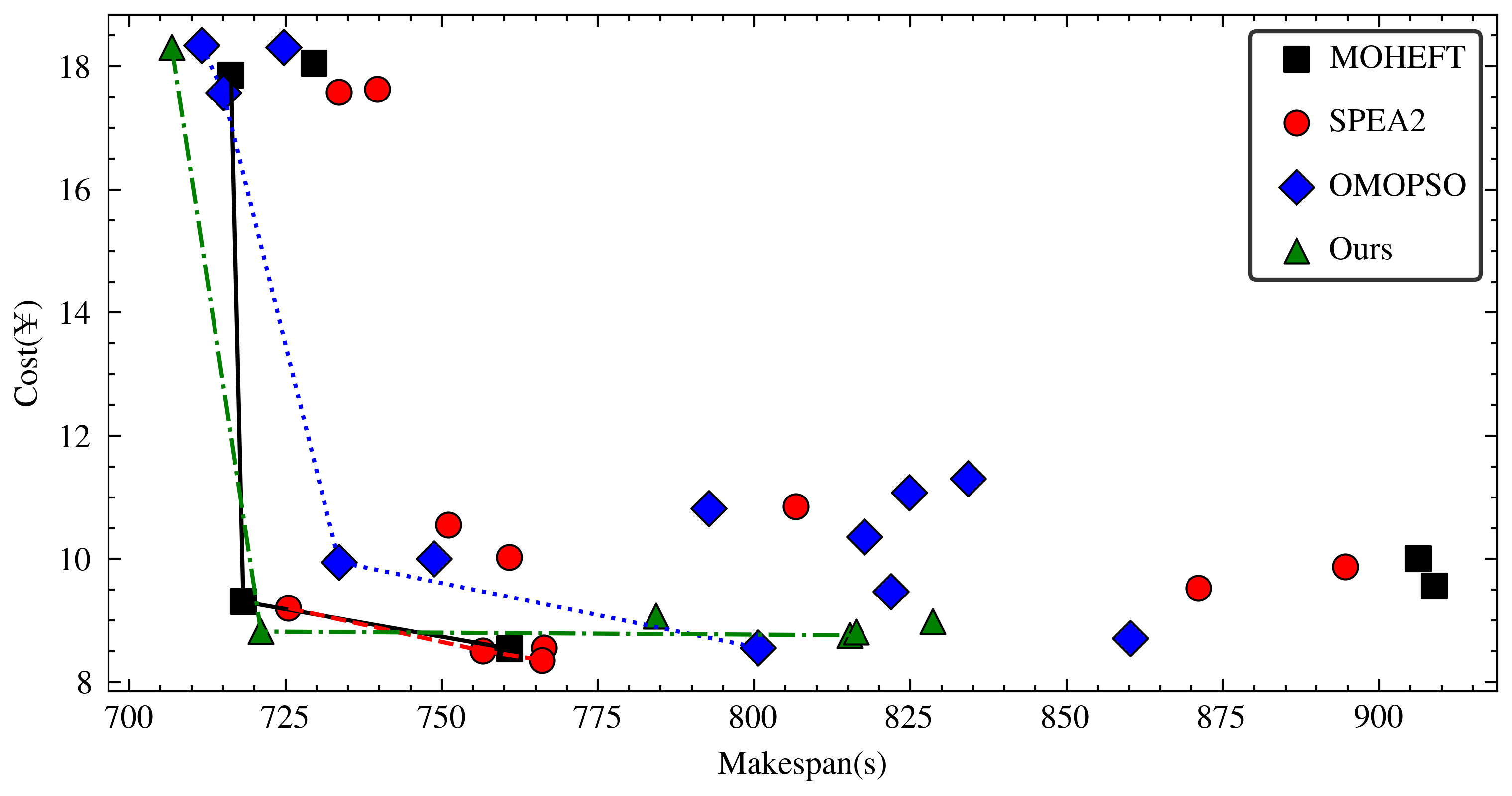}}
\subfloat[]{\includegraphics[width=2.35in]{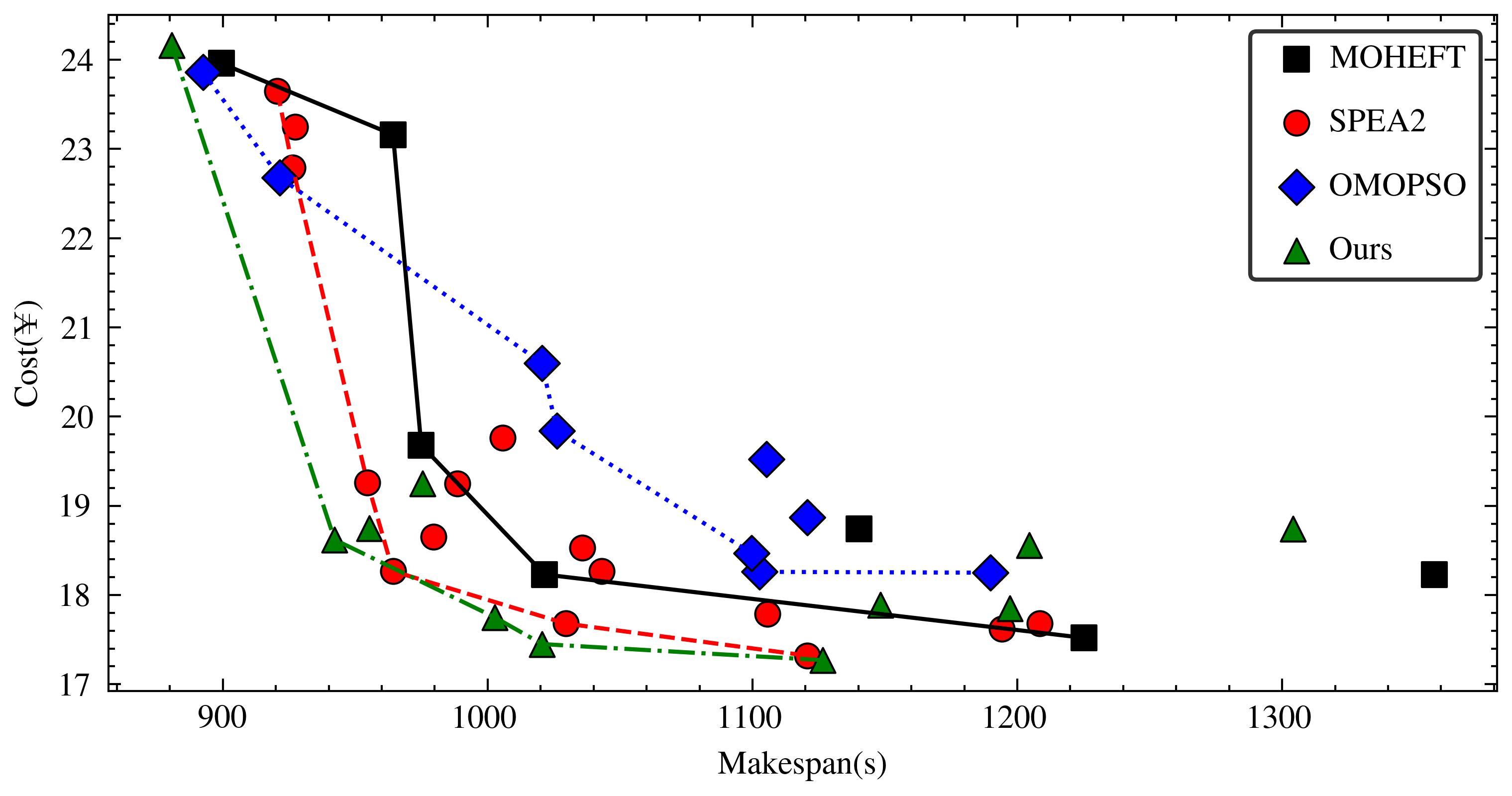}}
\quad
\subfloat[]{\includegraphics[width=2.35in]{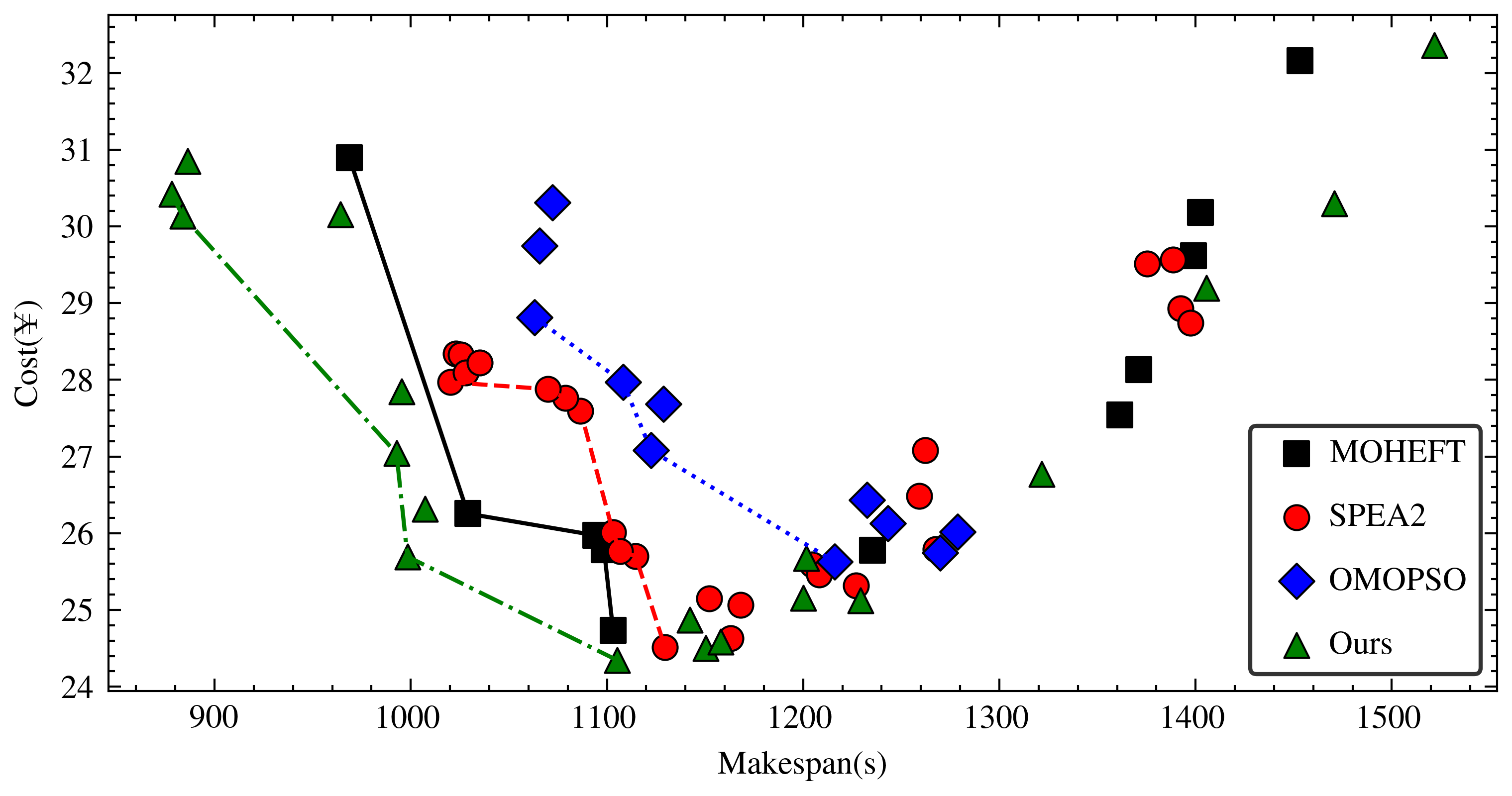}}
\quad
\subfloat[]{\includegraphics[width=2.35in]{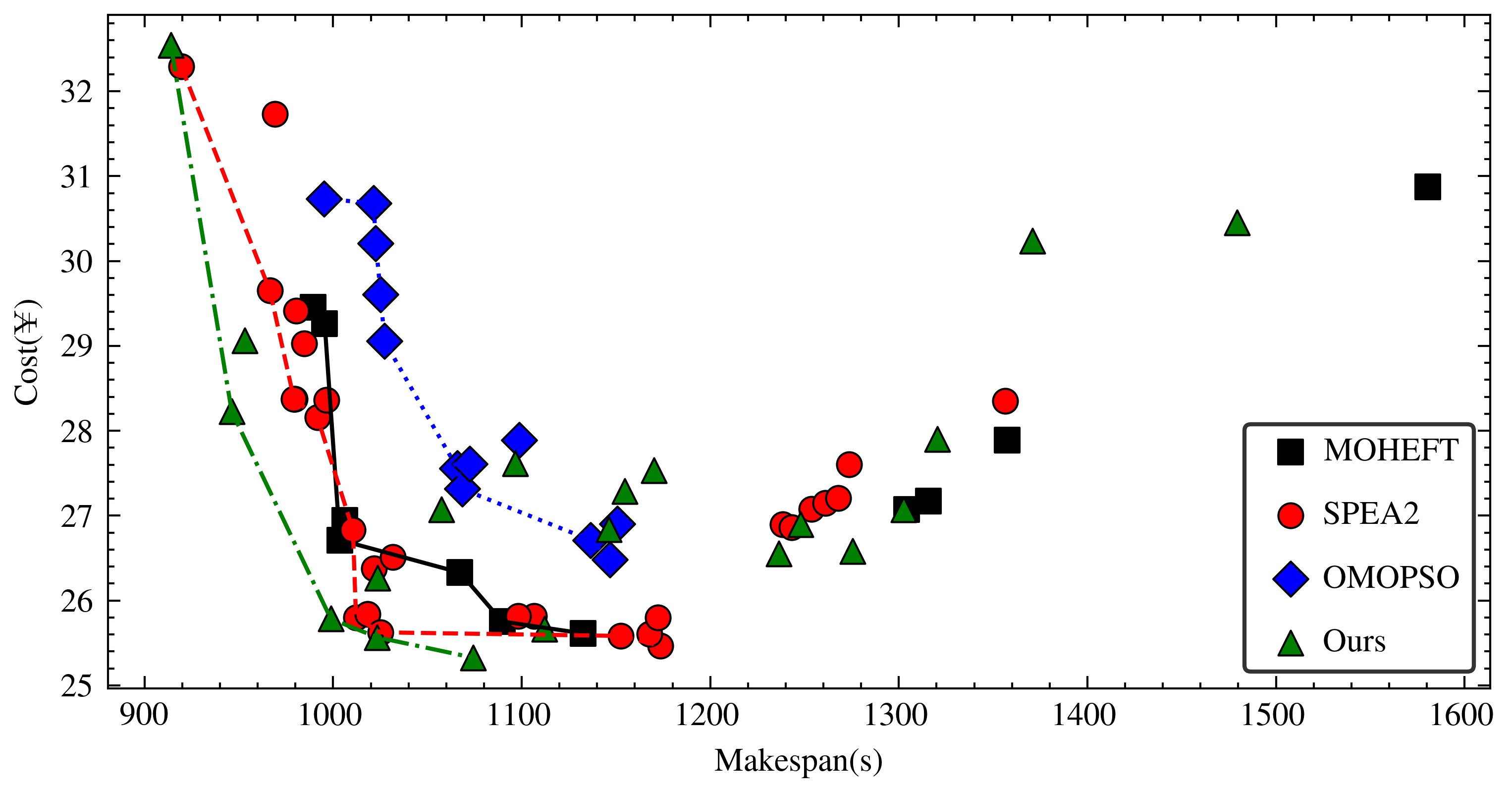}}
\caption{Monetery cost and makespan trade-offs experiments on real-world flowlines.
The dash in the lower left corner of each figure represents the \textit{Pareto Front} of the corresponding scheduling algorithm.
$x$m$y$o indicates the number of tasks $|V^m|$ and $|V^o|$ of the corresponding flowline.
(a) 3m6o. (b) 3m11o. (c) 4m8o. (d) 6m18o. (e) 6m29o.
}
\label{fig:exp}
\vspace{-0.6cm}
\end{figure*}

\Figure \ref{fig:exp} shows the results of scheduling experiments. Apparently, the results produced by our scheduling algorithm are significantly superior to others. Our scheduling results are leading the front of the MO optimization trade-off (closer to the lower-left corner in each subfigure) compared with the baselines.
That is, our algorithm for scheduling flowline can make a better solution in various scenarios that trade-off monetary cost with runtime for KGC.

The scatters tend to form multiple clusters when the number of tasks is small because different algorithms would produce the same scheduling combinations in the case of fewer tasks, resulting in close makespan and procurement costs. This trend disappears when the flowline includes more tasks because larger \emph{VM} clusters would introduce more randomness and perturbations such as network fluctuations.
Another trend is that our algorithm constructs a much smaller number of optional combinations than other algorithms. It benefits the GPU tasks guided compounding in Algorithm~\ref{alg:compound}, which reduces the search space for MO optimization.
In addition, we have marked the \emph{Pareto Front} of each scheduler with a dash in \Figure \ref{fig:exp}, which represents the set of Pareto efficient solutions, and it can be clearly observed that the \emph{Pareto Front} of our algorithm tends to outperform other MO schedulers.

These scheduling experiments demonstrate that our proposed scheduling algorithm can effectively schedule tasks in KGC (including IE models and operators, etc.) and outperforms the benchmark algorithms like MOHEFT, SPEA2, and OMOPSO in the scenario of KGC with multiple deep learning models by taking advantage of priori knowledge of KGC.

\subsubsection{Scaling Up} Once warmup is complete, the parameters and resource allocation are properly configured. In a distributed batch processing system like gBuilder, the following throughput would be stabilized. 
Therefore, the time cost of continuous KGC in  gBuilder scales nearly linearly with the data size, which is self-evident without further experiments.

\section{Conclusion}

This paper presents gBuilder, a high usability and scalability KGC system for extracting structured knowledge instances from the unstructured corpus by organizing multiple operators and IE models.
Compared with existing KGC systems and distributed computing systems, gBuilder combines their strengths, enabling flexible large-scale KGC through greedy graph partitioning scheduling algorithms over the proposed concept of \texttt{flowline}, which is an abstract of IE workflow and data pipeline.
Comprehensive end-to-end knowledge extraction and scheduling experiments demonstrate the usability and scalability of our proposed system. An extensive beta test\footnote{gBuilder is now avaliable at \url{http://gbuilder.gstore.cn}} and user feedback also show the usefulness of our system in KGC applications.

\bibliographystyle{IEEEtran}
\bibliography{ref}

\begin{thebibliography}{10}
\providecommand{\url}[1]{#1}
\csname url@samestyle\endcsname
\providecommand{\newblock}{\relax}
\providecommand{\bibinfo}[2]{#2}
\providecommand{\BIBentrySTDinterwordspacing}{\spaceskip=0pt\relax}
\providecommand{\BIBentryALTinterwordstretchfactor}{4}
\providecommand{\BIBentryALTinterwordspacing}{\spaceskip=\fontdimen2\font plus
\BIBentryALTinterwordstretchfactor\fontdimen3\font minus \fontdimen4\font\relax}
\providecommand{\BIBforeignlanguage}[2]{{%
\expandafter\ifx\csname l@#1\endcsname\relax
\typeout{** WARNING: IEEEtran.bst: No hyphenation pattern has been}%
\typeout{** loaded for the language `#1'. Using the pattern for}%
\typeout{** the default language instead.}%
\else
\language=\csname l@#1\endcsname
\fi
#2}}
\providecommand{\BIBdecl}{\relax}
\BIBdecl

\bibitem{Zou2013gStoreAG}
L.~Zou, M.~T. {\"O}zsu, L.~Chen, X.~Shen, R.~Huang, and D.~Zhao, ``gstore: a graph-based sparql query engine,'' \emph{The VLDB Journal}, vol.~23, pp. 565--590, 2013.

\bibitem{wilkinson2006jena}
K.~Wilkinson and K.~Wilkinson, ``Jena property table implementation,'' 2006.

\bibitem{zhang2015deepdive}
C.~Zhang, ``Deepdive: a data management system for automatic knowledge base construction,'' Ph.D. dissertation, The University of Wisconsin-Madison, 2015.

\bibitem{schneider2022decade}
P.~Schneider, T.~Schopf, J.~Vladika, M.~Galkin, E.~Simperl, and F.~Matthes, ``A decade of knowledge graphs in natural language processing: a survey,'' \emph{arXiv preprint arXiv:2210.00105}, 2022.

\bibitem{zou2020survey}
X.~Zou, ``A survey on application of knowledge graph,'' in \emph{Journal of Physics: Conference Series}, vol. 1487, no.~1.\hskip 1em plus 0.5em minus 0.4em\relax IOP Publishing, 2020, p. 012016.

\bibitem{wang2017knowledge}
Q.~Wang, Z.~Mao, B.~Wang, and L.~Guo, ``Knowledge graph embedding: A survey of approaches and applications,'' \emph{IEEE Transactions on Knowledge and Data Engineering}, vol.~29, no.~12, pp. 2724--2743, 2017.

\bibitem{nickel2015review}
M.~Nickel, K.~Murphy, V.~Tresp, and E.~Gabrilovich, ``A review of relational machine learning for knowledge graphs,'' \emph{Proceedings of the IEEE}, vol. 104, no.~1, pp. 11--33, 2015.

\bibitem{tiddi2022knowledge}
I.~Tiddi and S.~Schlobach, ``Knowledge graphs as tools for explainable machine learning: A survey,'' \emph{Artificial Intelligence}, vol. 302, p. 103627, 2022.

\bibitem{wang2023knowledgpt}
X.~Wang, Q.~Yang, Y.~Qiu, J.~Liang, Q.~He, Z.~Gu, Y.~Xiao, and W.~Wang, ``Knowledgpt: Enhancing large language models with retrieval and storage access on knowledge bases,'' \emph{arXiv preprint arXiv:2308.11761}, 2023.

\bibitem{feng2023trends}
Z.~Feng, W.~Ma, W.~Yu, L.~Huang, H.~Wang, Q.~Chen, W.~Peng, X.~Feng, B.~Qin \emph{et~al.}, ``Trends in integration of knowledge and large language models: A survey and taxonomy of methods, benchmarks, and applications,'' \emph{arXiv preprint arXiv:2311.05876}, 2023.

\bibitem{zhan2023admus}
Y.~Zhan, Y.~Li, M.~Zhang, and L.~Zou, ``Admus: A progressive question answering framework adaptable to multiple knowledge sources,'' \emph{arXiv preprint arXiv:2308.04800}, 2023.

\bibitem{auer2007dbpedia}
S.~Auer, C.~Bizer, G.~Kobilarov, J.~Lehmann, R.~Cyganiak, and Z.~Ives, ``Dbpedia: A nucleus for a web of open data,'' in \emph{The semantic web}.\hskip 1em plus 0.5em minus 0.4em\relax Springer, 2007, pp. 722--735.

\bibitem{bollacker2008freebase}
K.~Bollacker, C.~Evans, P.~Paritosh, T.~Sturge, and J.~Taylor, ``Freebase: a collaboratively created graph database for structuring human knowledge,'' in \emph{Proceedings of the 2008 ACM SIGMOD international conference on Management of data}, 2008, pp. 1247--1250.

\bibitem{hoffart2013yago2}
J.~Hoffart, F.~M. Suchanek, K.~Berberich, and G.~Weikum, ``Yago2: A spatially and temporally enhanced knowledge base from wikipedia,'' \emph{Artificial intelligence}, vol. 194, pp. 28--61, 2013.

\bibitem{abu2021domain}
B.~Abu-Salih, ``Domain-specific knowledge graphs: A survey,'' \emph{Journal of Network and Computer Applications}, vol. 185, p. 103076, 2021.

\bibitem{kondylakis2009ontology}
H.~Kondylakis, G.~Flouris, and D.~Plexousakis, ``Ontology and schema evolution in data integration: review and assessment,'' in \emph{On the Move to Meaningful Internet Systems: OTM 2009: Confederated International Conferences, CoopIS, DOA, IS, and ODBASE 2009, Vilamoura, Portugal, November 1-6, 2009, Proceedings, Part II}.\hskip 1em plus 0.5em minus 0.4em\relax Springer, 2009, pp. 932--947.

\bibitem{noy1997state}
N.~F. Noy and C.~D. Hafner, ``The state of the art in ontology design: A survey and comparative review,'' \emph{AI magazine}, vol.~18, no.~3, pp. 53--53, 1997.

\bibitem{nadeau2007survey}
D.~Nadeau and S.~Sekine, ``A survey of named entity recognition and classification,'' \emph{Lingvisticae Investigationes}, vol.~30, no.~1, pp. 3--26, 2007.

\bibitem{li2020survey}
J.~Li, A.~Sun, J.~Han, and C.~Li, ``A survey on deep learning for named entity recognition,'' \emph{IEEE Transactions on Knowledge and Data Engineering}, 2020.

\bibitem{kumar2017survey}
S.~Kumar, ``A survey of deep learning methods for relation extraction,'' \emph{arXiv preprint arXiv:1705.03645}, 2017.

\bibitem{de2006ontology}
J.~De~Bruijn, M.~Ehrig, C.~Feier, F.~Mart{\'\i}n-Recuerda, F.~Scharffe, and M.~Weiten, ``Ontology mediation, merging and aligning,'' \emph{Semantic web technologies}, pp. 95--113, 2006.

\bibitem{stumme2001ontology}
G.~Stumme and A.~Maedche, ``Ontology merging for federated ontologies on the semantic web.'' in \emph{OIS@ IJCAI}, 2001.

\bibitem{carley1988formalizing}
K.~Carley, ``Formalizing the social expert's knowledge,'' \emph{Sociological Methods \& Research}, vol.~17, no.~2, pp. 165--232, 1988.

\bibitem{kondreddi2014combining}
S.~K. Kondreddi, P.~Triantafillou, and G.~Weikum, ``Combining information extraction and human computing for crowdsourced knowledge acquisition,'' in \emph{2014 IEEE 30th International Conference on Data Engineering}.\hskip 1em plus 0.5em minus 0.4em\relax IEEE, 2014, pp. 988--999.

\bibitem{sarasua2015crowdsourcing}
C.~Sarasua, E.~Simperl, N.~F. Noy, A.~Bernstein, and J.~M. Leimeister, ``Crowdsourcing and the semantic web: A research manifesto,'' \emph{Human Computation}, vol.~2, no.~1, 2015.

\bibitem{paulheim2018much}
H.~Paulheim, ``How much is a triple?''\hskip 1em plus 0.5em minus 0.4em\relax ISWC, 2018.

\bibitem{Fensel2020}
\BIBentryALTinterwordspacing
D.~Fensel, U.~{\c{S}}im{\c{s}}ek, K.~Angele, E.~Huaman, E.~K{\"a}rle, O.~Panasiuk, I.~Toma, J.~Umbrich, and A.~Wahler, \emph{Introduction: What Is a Knowledge Graph?}\hskip 1em plus 0.5em minus 0.4em\relax Cham: Springer International Publishing, 2020, pp. 1--10. [Online]. Available: \url{https://doi.org/10.1007/978-3-030-37439-6_1}
\BIBentrySTDinterwordspacing

\bibitem{BERT}
\BIBentryALTinterwordspacing
J.~Devlin, M.-W. Chang, K.~Lee, and K.~Toutanova, ``{BERT}: Pre-training of deep bidirectional transformers for language understanding,'' in \emph{Proceedings of the 2019 Conference of the North {A}merican Chapter of the Association for Computational Linguistics: Human Language Technologies, Volume 1 (Long and Short Papers)}.\hskip 1em plus 0.5em minus 0.4em\relax Minneapolis, Minnesota: Association for Computational Linguistics, 2019, pp. 4171--4186. [Online]. Available: \url{https://www.aclweb.org/anthology/N19-1423}
\BIBentrySTDinterwordspacing

\bibitem{GPT}
A.~Radford, K.~Narasimhan, T.~Salimans, and I.~Sutskever, ``Improving language understanding with unsupervised learning,'' 2018.

\bibitem{XLNet}
\BIBentryALTinterwordspacing
Z.~Yang, Z.~Dai, Y.~Yang, J.~G. Carbonell, R.~Salakhutdinov, and Q.~V. Le, ``Xlnet: Generalized autoregressive pretraining for language understanding,'' in \emph{Advances in Neural Information Processing Systems 32: Annual Conference on Neural Information Processing Systems 2019, NeurIPS 2019, December 8-14, 2019, Vancouver, BC, Canada}, H.~M. Wallach, H.~Larochelle, A.~Beygelzimer, F.~d'Alch{\'{e}}{-}Buc, E.~B. Fox, and R.~Garnett, Eds., 2019, pp. 5754--5764. [Online]. Available: \url{https://proceedings.neurips.cc/paper/2019/hash/dc6a7e655d7e5840e66733e9ee67cc69-Abstract.html}
\BIBentrySTDinterwordspacing

\bibitem{openai2023gpt4}
OpenAI, ``Gpt-4 technical report,'' 2023.

\bibitem{touvron2023llama}
H.~Touvron, T.~Lavril, G.~Izacard, X.~Martinet, M.-A. Lachaux, T.~Lacroix, B.~Rozi{\`e}re, N.~Goyal, E.~Hambro, F.~Azhar \emph{et~al.}, ``Llama: Open and efficient foundation language models,'' \emph{arXiv preprint arXiv:2302.13971}, 2023.

\bibitem{wei2023zero}
X.~Wei, X.~Cui, N.~Cheng, X.~Wang, X.~Zhang, S.~Huang, P.~Xie, J.~Xu, Y.~Chen, M.~Zhang \emph{et~al.}, ``Zero-shot information extraction via chatting with chatgpt,'' \emph{arXiv preprint arXiv:2302.10205}, 2023.

\bibitem{gupta2022matscibert}
T.~Gupta, M.~Zaki, N.~A. Krishnan, and Mausam, ``Matscibert: A materials domain language model for text mining and information extraction,'' \emph{npj Computational Materials}, vol.~8, no.~1, p. 102, 2022.

\bibitem{zhang2023llmaaa}
R.~Zhang, Y.~Li, Y.~Ma, M.~Zhou, and L.~Zou, ``Llmaaa: Making large language models as active annotators,'' \emph{arXiv preprint arXiv:2310.19596}, 2023.

\bibitem{arxiv.2201.03335}
\BIBentryALTinterwordspacing
N.~Zhang, X.~Xu, L.~Tao, H.~Yu, H.~Ye, X.~Xie, X.~Chen, Z.~Li, L.~Li, X.~Liang, Y.~Yao, S.~Deng, W.~Zhang, Z.~Zhang, C.~Tan, F.~Huang, G.~Zheng, and H.~Chen, ``Deepke: A deep learning based knowledge extraction toolkit for knowledge base population,'' 2022. [Online]. Available: \url{https://arxiv.org/abs/2201.03335}
\BIBentrySTDinterwordspacing

\bibitem{kertkeidkachorn2017t2kg}
N.~Kertkeidkachorn and R.~Ichise, ``T2kg: An end-to-end system for creating knowledge graph from unstructured text,'' in \emph{Workshops at the Thirty-First AAAI Conference on Artificial Intelligence}, 2017.

\bibitem{li-etal-2021-tdeer}
\BIBentryALTinterwordspacing
X.~Li, X.~Luo, C.~Dong, D.~Yang, B.~Luan, and Z.~He, ``{TDEER}: An efficient translating decoding schema for joint extraction of entities and relations,'' in \emph{Proceedings of the 2021 Conference on Empirical Methods in Natural Language Processing}.\hskip 1em plus 0.5em minus 0.4em\relax Online and Punta Cana, Dominican Republic: Association for Computational Linguistics, Nov. 2021, pp. 8055--8064. [Online]. Available: \url{https://aclanthology.org/2021.emnlp-main.635}
\BIBentrySTDinterwordspacing

\bibitem{sui2020joint}
D.~Sui, Y.~Chen, K.~Liu, J.~Zhao, X.~Zeng, and S.~Liu, ``Joint entity and relation extraction with set prediction networks,'' \emph{arXiv preprint arXiv:2011.01675}, 2020.

\bibitem{wang2022deepstruct}
C.~Wang, X.~Liu, Z.~Chen, H.~Hong, J.~Tang, and D.~Song, ``Deepstruct: Pretraining of language models for structure prediction,'' \emph{arXiv preprint arXiv:2205.10475}, 2022.

\bibitem{kejriwal2019domain}
M.~Kejriwal, \emph{Domain-specific knowledge graph construction}.\hskip 1em plus 0.5em minus 0.4em\relax Springer, 2019.

\bibitem{wang2018information}
C.~Wang, X.~Ma, J.~Chen, and J.~Chen, ``Information extraction and knowledge graph construction from geoscience literature,'' \emph{Computers \& geosciences}, vol. 112, pp. 112--120, 2018.

\bibitem{anantharangachar2013ontology}
R.~Anantharangachar, S.~Ramani, and S.~Rajagopalan, ``Ontology guided information extraction from unstructured text,'' \emph{arXiv preprint arXiv:1302.1335}, 2013.

\bibitem{musen2015protege}
M.~A. Musen, ``The prot{\'e}g{\'e} project: a look back and a look forward,'' \emph{AI matters}, vol.~1, no.~4, pp. 4--12, 2015.

\bibitem{chen2020review}
X.~Chen, S.~Jia, and Y.~Xiang, ``A review: Knowledge reasoning over knowledge graph,'' \emph{Expert Systems with Applications}, vol. 141, p. 112948, 2020.

\bibitem{zhao2018architecture}
Z.~Zhao, S.-K. Han, and I.-M. So, ``Architecture of knowledge graph construction techniques,'' \emph{International Journal of Pure and Applied Mathematics}, vol. 118, no.~19, pp. 1869--1883, 2018.

\bibitem{lee2007automated}
C.-S. Lee, Y.-F. Kao, Y.-H. Kuo, and M.-H. Wang, ``Automated ontology construction for unstructured text documents,'' \emph{Data \& Knowledge Engineering}, vol.~60, no.~3, pp. 547--566, 2007.

\bibitem{10.1145/3583780.3615178}
\BIBentryALTinterwordspacing
Y.~Li, S.~Hu, W.~Han, and L.~Zou, ``Cord: A three-stage coarse-to-fine framework for relation detection in knowledge base question answering,'' in \emph{Proceedings of the 32nd ACM International Conference on Information and Knowledge Management}, ser. CIKM '23.\hskip 1em plus 0.5em minus 0.4em\relax New York, NY, USA: Association for Computing Machinery, 2023, p. 4069–4073. [Online]. Available: \url{https://doi.org/10.1145/3583780.3615178}
\BIBentrySTDinterwordspacing

\bibitem{wei2019novel}
\BIBentryALTinterwordspacing
Z.~Wei, J.~Su, Y.~Wang, Y.~Tian, and Y.~Chang, ``A novel cascade binary tagging framework for relational triple extraction,'' in \emph{Proceedings of the 58th Annual Meeting of the Association for Computational Linguistics}.\hskip 1em plus 0.5em minus 0.4em\relax Online: Association for Computational Linguistics, 2020, pp. 1476--1488. [Online]. Available: \url{https://www.aclweb.org/anthology/2020.acl-main.136}
\BIBentrySTDinterwordspacing

\bibitem{soares2019matching}
\BIBentryALTinterwordspacing
L.~Baldini~Soares, N.~FitzGerald, J.~Ling, and T.~Kwiatkowski, ``Matching the blanks: Distributional similarity for relation learning,'' in \emph{Proceedings of the 57th Annual Meeting of the Association for Computational Linguistics}.\hskip 1em plus 0.5em minus 0.4em\relax Florence, Italy: Association for Computational Linguistics, 2019, pp. 2895--2905. [Online]. Available: \url{https://www.aclweb.org/anthology/P19-1279}
\BIBentrySTDinterwordspacing

\bibitem{sun2019distantly}
C.~Sun and Y.~Wu, ``Distantly supervised entity relation extraction with adapted manual annotations,'' in \emph{Proceedings of the AAAI conference on artificial intelligence}, vol.~33, no.~01, 2019, pp. 7039--7046.

\bibitem{zhang2021semi}
L.~Zhang, Y.~Li, R.~Zhang, and W.~Li, ``Semi-open attribute extraction from chinese functional description text,'' in \emph{Asian Conference on Machine Learning}.\hskip 1em plus 0.5em minus 0.4em\relax PMLR, 2021, pp. 1505--1520.

\bibitem{li-etal-2023-attgen}
\BIBentryALTinterwordspacing
Y.~Li, B.~Xue, R.~Zhang, and L.~Zou, ``{A}t{TG}en: Attribute tree generation for real-world attribute joint extraction,'' in \emph{Proceedings of the 61st Annual Meeting of the Association for Computational Linguistics (Volume 1: Long Papers)}.\hskip 1em plus 0.5em minus 0.4em\relax Toronto, Canada: Association for Computational Linguistics, Jul. 2023, pp. 2139--2152. [Online]. Available: \url{https://aclanthology.org/2023.acl-long.119}
\BIBentrySTDinterwordspacing

\bibitem{vandic2012faceted}
D.~Vandic, J.-W. Van~Dam, and F.~Frasincar, ``Faceted product search powered by the semantic web,'' \emph{Decision Support Systems}, vol.~53, no.~3, pp. 425--437, 2012.

\bibitem{more2016attribute}
A.~More, ``Attribute extraction from product titles in ecommerce,'' \emph{arXiv preprint arXiv:1608.04670}, 2016.

\bibitem{wang2023gptner}
S.~Wang, X.~Sun, X.~Li, R.~Ouyang, F.~Wu, T.~Zhang, J.~Li, and G.~Wang, ``Gpt-ner: Named entity recognition via large language models,'' 2023.

\bibitem{zhou2021ensemble}
Z.-H. Zhou, ``Ensemble learning,'' in \emph{Machine learning}.\hskip 1em plus 0.5em minus 0.4em\relax Springer, 2021, pp. 181--210.

\bibitem{stuckenschmidt2002approximate}
H.~Stuckenschmidt, ``Approximate information filtering on the semantic web,'' in \emph{Annual Conference on Artificial Intelligence}.\hskip 1em plus 0.5em minus 0.4em\relax Springer, 2002, pp. 114--128.

\bibitem{hitzler2005ontology}
P.~Hitzler, M.~Krotzsch, M.~Ehrig, and Y.~Sure, ``What is ontology merging?--a categorytheoretical perspective using pushouts. american association of artificial intelligence,'' 2005.

\bibitem{kalfoglou2002information}
Y.~Kalfoglou and M.~Schorlemmer, ``Information-flow-based ontology mapping,'' in \emph{OTM Confederated International Conferences" On the Move to Meaningful Internet Systems"}.\hskip 1em plus 0.5em minus 0.4em\relax Springer, 2002, pp. 1132--1151.

\bibitem{paccanaro2001learning}
A.~Paccanaro and G.~E. Hinton, ``Learning distributed representations of concepts using linear relational embedding,'' \emph{IEEE Transactions on Knowledge and Data Engineering}, vol.~13, no.~2, pp. 232--244, 2001.

\bibitem{celikyilmaz2015investigation}
A.~Celikyilmaz and D.~Hakkani-Tur, ``Investigation of ensemble models for sequence learning,'' in \emph{2015 IEEE International Conference on Acoustics, Speech and Signal Processing (ICASSP)}.\hskip 1em plus 0.5em minus 0.4em\relax IEEE, 2015, pp. 5381--5385.

\bibitem{warren2015big}
J.~Warren and N.~Marz, \emph{Big Data: Principles and best practices of scalable realtime data systems}.\hskip 1em plus 0.5em minus 0.4em\relax Simon and Schuster, 2015.

\bibitem{spark}
M.~Zaharia, R.~S. Xin, P.~Wendell, T.~Das, M.~Armbrust, A.~Dave, X.~Meng, J.~Rosen, S.~Venkataraman, M.~J. Franklin \emph{et~al.}, ``Apache spark: a unified engine for big data processing,'' \emph{Communications of the ACM}, vol.~59, no.~11, pp. 56--65, 2016.

\bibitem{cudnn}
S.~Chetlur, C.~Woolley, P.~Vandermersch, J.~Cohen, J.~Tran, B.~Catanzaro, and E.~Shelhamer, ``cudnn: Efficient primitives for deep learning,'' \emph{arXiv preprint arXiv:1410.0759}, 2014.

\bibitem{portella2019statistical}
G.~Portella, G.~N. Rodrigues, E.~Nakano, and A.~C. Melo, ``Statistical analysis of amazon ec2 cloud pricing models,'' \emph{Concurrency and Computation: Practice and Experience}, vol.~31, no.~18, p. e4451, 2019.

\bibitem{khan2022exploration}
M.~Khan, A.~I. Jehangiri, Z.~Ahmad, M.~A. Ala’anzy, and A.~Umer, ``An exploration to graphics processing unit spot price prediction,'' \emph{Cluster Computing}, pp. 1--17, 2022.

\bibitem{bukh1992art}
P.~N.~D. Bukh, ``The art of computer systems performance analysis, techniques for experimental design, measurement, simulation and modeling,'' 1992.

\bibitem{su2013cost}
S.~Su, J.~Li, Q.~Huang, X.~Huang, K.~Shuang, and J.~Wang, ``Cost-efficient task scheduling for executing large programs in the cloud,'' \emph{Parallel Computing}, vol.~39, no. 4-5, pp. 177--188, 2013.

\bibitem{kurowski2006grid}
K.~Kurowski, J.~Nabrzyski, A.~Oleksiak, and J.~Weglarz, ``Grid multicriteria job scheduling with resource reservation and prediction mechanisms,'' in \emph{Perspectives in modern project scheduling}.\hskip 1em plus 0.5em minus 0.4em\relax Springer, 2006, pp. 345--373.

\bibitem{wei2020CasRel}
Z.~Wei, J.~Su, Y.~Wang, Y.~Tian, and Y.~Chang, ``A novel cascade binary tagging framework for relational triple extraction,'' in \emph{Proceedings of the 58th Annual Meeting of the Association for Computational Linguistics}, 2020, pp. 1476--1488.

\bibitem{sandhaus2008new}
E.~Sandhaus, ``The new york times annotated corpus,'' \emph{Linguistic Data Consortium, Philadelphia}, vol.~6, no.~12, p. e26752, 2008.

\bibitem{zeng2018extracting}
X.~Zeng, D.~Zeng, S.~He, K.~Liu, and J.~Zhao, ``Extracting relational facts by an end-to-end neural model with copy mechanism,'' in \emph{Proceedings of the 56th Annual Meeting of the Association for Computational Linguistics (Volume 1: Long Papers)}, 2018, pp. 506--514.

\bibitem{web_nlg}
\BIBentryALTinterwordspacing
C.~Gardent, A.~Shimorina, S.~Narayan, and L.~Perez{-}Beltrachini, ``Creating training corpora for {NLG} micro-planners,'' in \emph{Proceedings of the 55th Annual Meeting of the Association for Computational Linguistics, {ACL} 2017, Vancouver, Canada, July 30 - August 4, Volume 1: Long Papers}, R.~Barzilay and M.~Kan, Eds.\hskip 1em plus 0.5em minus 0.4em\relax Association for Computational Linguistics, 2017, pp. 179--188. [Online]. Available: \url{https://doi.org/10.18653/v1/P17-1017}
\BIBentrySTDinterwordspacing

\bibitem{zhang2017tacred}
\BIBentryALTinterwordspacing
Y.~Zhang, V.~Zhong, D.~Chen, G.~Angeli, and C.~D. Manning, ``Position-aware attention and supervised data improve slot filling,'' in \emph{Proceedings of the 2017 Conference on Empirical Methods in Natural Language Processing (EMNLP 2017)}, 2017, pp. 35--45. [Online]. Available: \url{https://nlp.stanford.edu/pubs/zhang2017tacred.pdf}
\BIBentrySTDinterwordspacing

\bibitem{li2019duie}
S.~Li, W.~He, Y.~Shi, W.~Jiang, H.~Liang, Y.~Jiang, Y.~Zhang, Y.~Lyu, and Y.~Zhu, ``Duie: A large-scale chinese dataset for information extraction,'' in \emph{CCF International Conference on Natural Language Processing and Chinese Computing}.\hskip 1em plus 0.5em minus 0.4em\relax Springer, 2019, pp. 791--800.

\bibitem{agrawal-etal-2022-large}
\BIBentryALTinterwordspacing
M.~Agrawal, S.~Hegselmann, H.~Lang, Y.~Kim, and D.~Sontag, ``Large language models are few-shot clinical information extractors,'' in \emph{Proceedings of the 2022 Conference on Empirical Methods in Natural Language Processing}.\hskip 1em plus 0.5em minus 0.4em\relax Abu Dhabi, United Arab Emirates: Association for Computational Linguistics, Dec. 2022, pp. 1998--2022. [Online]. Available: \url{https://aclanthology.org/2022.emnlp-main.130}
\BIBentrySTDinterwordspacing

\bibitem{qiu2020pre}
X.~Qiu, T.~Sun, Y.~Xu, Y.~Shao, N.~Dai, and X.~Huang, ``Pre-trained models for natural language processing: A survey,'' \emph{Science China Technological Sciences}, vol.~63, no.~10, pp. 1872--1897, 2020.

\bibitem{li-etal-2022-enhancing}
\BIBentryALTinterwordspacing
Y.~Li, J.~Cao, X.~Cong, Z.~Zhang, B.~Yu, H.~Zhu, and T.~Liu, ``Enhancing {C}hinese pre-trained language model via heterogeneous linguistics graph,'' in \emph{Proceedings of the 60th Annual Meeting of the Association for Computational Linguistics (Volume 1: Long Papers)}.\hskip 1em plus 0.5em minus 0.4em\relax Dublin, Ireland: Association for Computational Linguistics, May 2022, pp. 1986--1996. [Online]. Available: \url{https://aclanthology.org/2022.acl-long.140}
\BIBentrySTDinterwordspacing

\bibitem{zhang2014multi}
F.~Zhang, J.~Cao, K.~Li, S.~U. Khan, and K.~Hwang, ``Multi-objective scheduling of many tasks in cloud platforms,'' \emph{Future Generation Computer Systems}, vol.~37, pp. 309--320, 2014.

\bibitem{durillo2014multi}
J.~J. Durillo and R.~Prodan, ``Multi-objective workflow scheduling in amazon ec2,'' \emph{Cluster computing}, vol.~17, no.~2, pp. 169--189, 2014.

\bibitem{zhou2019minimizing}
X.~Zhou, G.~Zhang, J.~Sun, J.~Zhou, T.~Wei, and S.~Hu, ``Minimizing cost and makespan for workflow scheduling in cloud using fuzzy dominance sort based heft,'' \emph{Future Generation Computer Systems}, vol.~93, pp. 278--289, 2019.

\bibitem{Durillo2012MOHEFTAM}
J.~J. Durillo, H.~M. Fard, and R.~Prodan, ``Moheft: A multi-objective list-based method for workflow scheduling,'' \emph{4th IEEE International Conference on Cloud Computing Technology and Science Proceedings}, pp. 185--192, 2012.

\bibitem{Zitzler2002SPEA2IT}
E.~Zitzler, M.~Laumanns, and L.~Thiele, ``Spea2: Improving the strength pareto evolutionary algorithm for multiobjective optimization,'' 2002.

\bibitem{yu2007multi}
J.~Yu, M.~Kirley, and R.~Buyya, ``Multi-objective planning for workflow execution on grids,'' in \emph{2007 8th IEEE/ACM International Conference on Grid Computing}.\hskip 1em plus 0.5em minus 0.4em\relax IEEE, 2007, pp. 10--17.

\bibitem{RC05}
M.~Reyes and C.~{Coello Coello}, ``Improving pso-based multi-objective optimization using crowding, mutation and $\epsilon$-dominance,'' in \emph{Third International Conference on Evolutionary Multi\-Criterion Optimization, EMO 2005}, ser. LNCS, C.~Coello, A.~Hern\'{a}ndez, and E.~Zitler, Eds., vol. 3410.\hskip 1em plus 0.5em minus 0.4em\relax Springer, 2005, pp. 509--519.

\bibitem{kaur2018novel}
M.~Kaur and S.~Kadam, ``A novel multi-objective bacteria foraging optimization algorithm (mobfoa) for multi-objective scheduling,'' \emph{Applied Soft Computing}, vol.~66, pp. 183--195, 2018.

\bibitem{BENITEZHIDALGO2019100598}
\BIBentryALTinterwordspacing
A.~Benítez-Hidalgo, A.~J. Nebro, J.~García-Nieto, I.~Oregi, and J.~D. Ser, ``jmetalpy: A python framework for multi-objective optimization with metaheuristics,'' \emph{Swarm and Evolutionary Computation}, p. 100598, 2019. [Online]. Available: \url{http://www.sciencedirect.com/science/article/pii/S2210650219301397}
\BIBentrySTDinterwordspacing

\bibitem{geng-etal-2021-fasthan}
\BIBentryALTinterwordspacing
Z.~Geng, H.~Yan, X.~Qiu, and X.~Huang, ``fasthan: A bert-based multi-task toolkit for chinese nlp,'' in \emph{Proceedings of the 59th Annual Meeting of the Association for Computational Linguistics and the 11th International Joint Conference on Natural Language Processing: System Demonstrations}, 2021, pp. 99--106. [Online]. Available: \url{https://aclanthology.org/2021.acl-demo.12}
\BIBentrySTDinterwordspacing

\bibitem{li-etal-2020-flat}
\BIBentryALTinterwordspacing
X.~Li, H.~Yan, X.~Qiu, and X.~Huang, ``{FLAT}: {C}hinese {NER} using flat-lattice transformer,'' in \emph{Proceedings of the 58th Annual Meeting of the Association for Computational Linguistics}.\hskip 1em plus 0.5em minus 0.4em\relax Online: Association for Computational Linguistics, Jul. 2020, pp. 6836--6842. [Online]. Available: \url{https://aclanthology.org/2020.acl-main.611}
\BIBentrySTDinterwordspacing

\bibitem{baldini-soares-etal-2019-matching}
\BIBentryALTinterwordspacing
L.~Baldini~Soares, N.~FitzGerald, J.~Ling, and T.~Kwiatkowski, ``Matching the blanks: Distributional similarity for relation learning,'' in \emph{Proceedings of the 57th Annual Meeting of the Association for Computational Linguistics}.\hskip 1em plus 0.5em minus 0.4em\relax Florence, Italy: Association for Computational Linguistics, Jul. 2019, pp. 2895--2905. [Online]. Available: \url{https://aclanthology.org/P19-1279}
\BIBentrySTDinterwordspacing

\bibitem{cui-etal-2021-pretrain}
\BIBentryALTinterwordspacing
Y.~Cui, W.~Che, T.~Liu, B.~Qin, and Z.~Yang, ``Pre-training with whole word masking for chinese bert,'' 2021. [Online]. Available: \url{https://ieeexplore.ieee.org/document/9599397}
\BIBentrySTDinterwordspacing

\bibitem{wang-etal-2020-tplinker}
\BIBentryALTinterwordspacing
Y.~Wang, B.~Yu, Y.~Zhang, T.~Liu, H.~Zhu, and L.~Sun, ``{TPL}inker: Single-stage joint extraction of entities and relations through token pair linking,'' in \emph{Proceedings of the 28th International Conference on Computational Linguistics}.\hskip 1em plus 0.5em minus 0.4em\relax Barcelona, Spain (Online): International Committee on Computational Linguistics, Dec. 2020, pp. 1572--1582. [Online]. Available: \url{https://www.aclweb.org/anthology/2020.coling-main.138}
\BIBentrySTDinterwordspacing

\bibitem{tjong-kim-sang-de-meulder-2003-introduction}
\BIBentryALTinterwordspacing
E.~F. Tjong Kim~Sang and F.~De~Meulder, ``Introduction to the {C}o{NLL}-2003 shared task: Language-independent named entity recognition,'' in \emph{Proceedings of the Seventh Conference on Natural Language Learning at {HLT}-{NAACL} 2003}, 2003, pp. 142--147. [Online]. Available: \url{https://aclanthology.org/W03-0419}
\BIBentrySTDinterwordspacing

\bibitem{wang2019ipre}
H.~Wang, Z.~He, J.~Ma, W.~Chen, and M.~Zhang, ``Ipre: a dataset for inter-personal relationship extraction,'' in \emph{CCF International Conference on Natural Language Processing and Chinese Computing}.\hskip 1em plus 0.5em minus 0.4em\relax Springer, 2019, pp. 103--115.

\bibitem{zhang-yang-2018-chinese}
\BIBentryALTinterwordspacing
Y.~Zhang and J.~Yang, ``{C}hinese {NER} using lattice {LSTM},'' in \emph{Proceedings of the 56th Annual Meeting of the Association for Computational Linguistics (Volume 1: Long Papers)}.\hskip 1em plus 0.5em minus 0.4em\relax Melbourne, Australia: Association for Computational Linguistics, Jul. 2018, pp. 1554--1564. [Online]. Available: \url{https://www.aclweb.org/anthology/P18-1144}
\BIBentrySTDinterwordspacing

\bibitem{garey1979computers}
M.~R. Garey and D.~S. Johnson, \emph{Computers and intractability}.\hskip 1em plus 0.5em minus 0.4em\relax freeman San Francisco, 1979, vol. 174.

\bibitem{10.1145/361011.361064}
\BIBentryALTinterwordspacing
J.~Bruno, E.~G. Coffman, and R.~Sethi, ``Scheduling independent tasks to reduce mean finishing time,'' \emph{Commun. ACM}, vol.~17, no.~7, p. 382–387, jul 1974. [Online]. Available: \url{https://doi.org/10.1145/361011.361064}
\BIBentrySTDinterwordspacing

\end{thebibliography}


\clearpage
\appendices

\section{System Architecture}\label{app:system}

gBuilder is proposed to construct KGs from large-scale unstructured text. It is modularity developed following the microservices architecture and supports concurrent KGC for multiple projects and multi-users. Furthermore, gBuilder deploys in multi-cloud environments to scale-up computing resources at cloud providers to dynamically meet the requirements of large-scale KGC. This section introduces the architecture of gBuilder functionally. 
\begin{figure}[h]
\centering{\includegraphics[width=.9\linewidth]{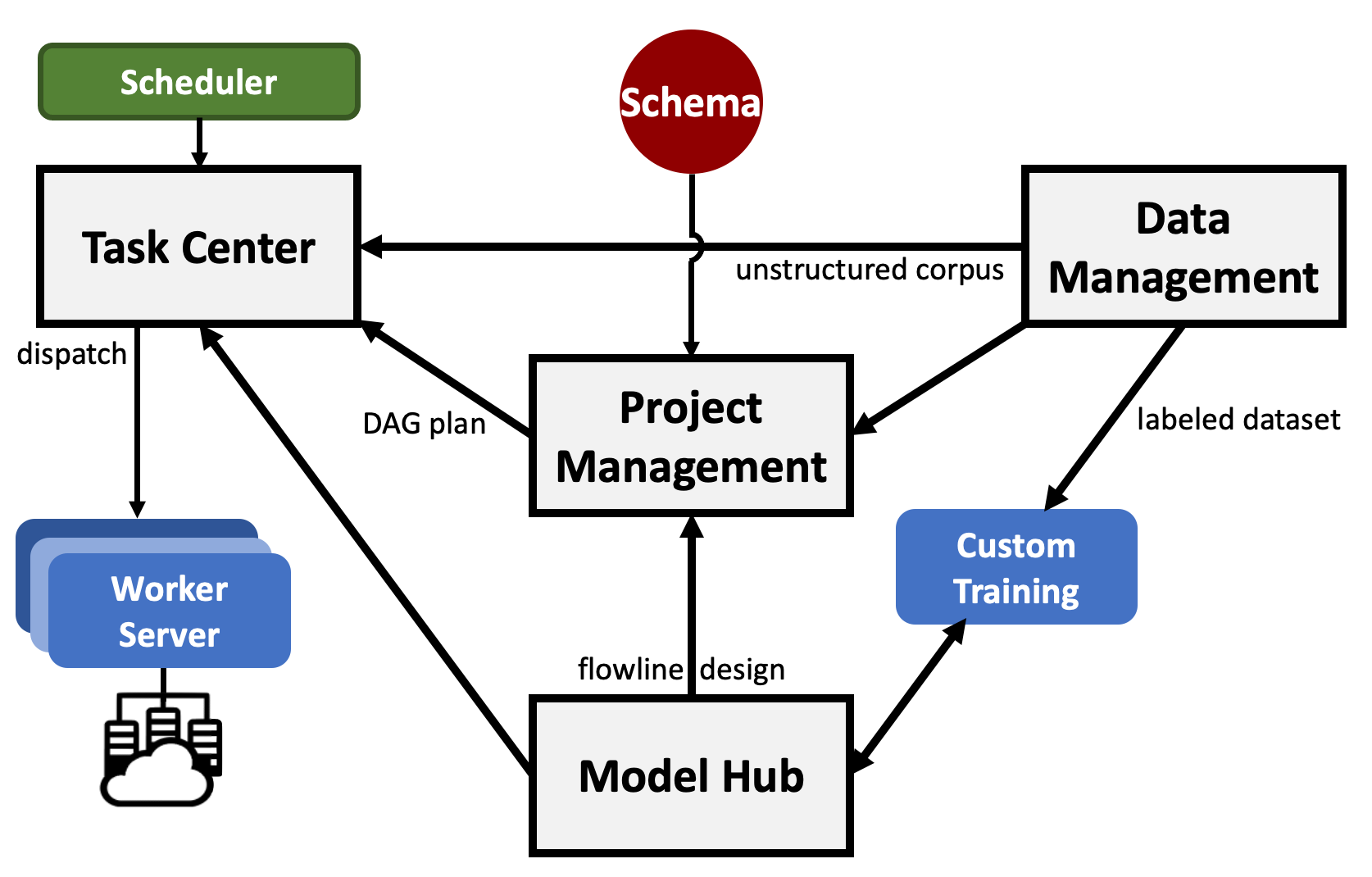}}
	\caption{The architecture of gBuilder.}
	\label{fig:architecture}
\end{figure}


As \Figure \ref{fig:architecture} shows, gBuilder is composed of four main modules.

\textbf{Project Management} module is the core component of gBuilder. 
It controls the entire KBC process, including Flowline-related features (design, storage, verify), user authentication, logging collector, etc. This module also schedules other modules through inter-services communication and manages the KGC workflow from import of schema to generation of DAG execution plan.

\textbf{Data Management} module manages the corpora of the project, including unstructured corpus for inference and labeled data for training. This module is the data source of KGC. It will divide the data by column or row according to the needs of the task center and push the data slice to the specified service.

\textbf{Model Hub} manages all deep learning models in gBuilder, including built-in models and custom models. When a task calls for a well-tuned IE model, the model hub will transfer the skeleton and weights of a model to the corresponding worker node and instantiate it on the worker server. In addition, the model hub also controls the training process of custom models, including applying computing resources, controlling hyperparameters tuning, etc. We also provide an interface to accept user-custom model endpoints to contain users' self-server models or algorithms in Model Hub and insert them into the KGC flowline.

\textbf{Task Center} controls the scheduling of tasks. This module will further divide the DAG execution plan obtained from the project management module and find out the parallel and serial relation in the DAG.
The isolative sequence of tasks that can be executed independently will be integrated into one job. Then the task center will apply and assign computing resources on demand for each job. The operator and task-related data corresponding to the job will be instantiated on the terminal node. Besides, the task center also collects generated loggings, and monitors the execution status of each node and the transmission of data.

Besides the four main modules of the core service, \textbf{Worker Server} is also an essential part of gBuilder. It is the service that instantiates on the terminal computing node, executes the job according to the schedule from the task center, instantiates the IE model transmitted from the Model Hub, obtains input data from the predecessor tasks, and preprocesses output data for the successor tasks.

\section{User Interfaces}

\begin{figure}[h]
\centering
\subfloat[]{\includegraphics[width=3.35in]{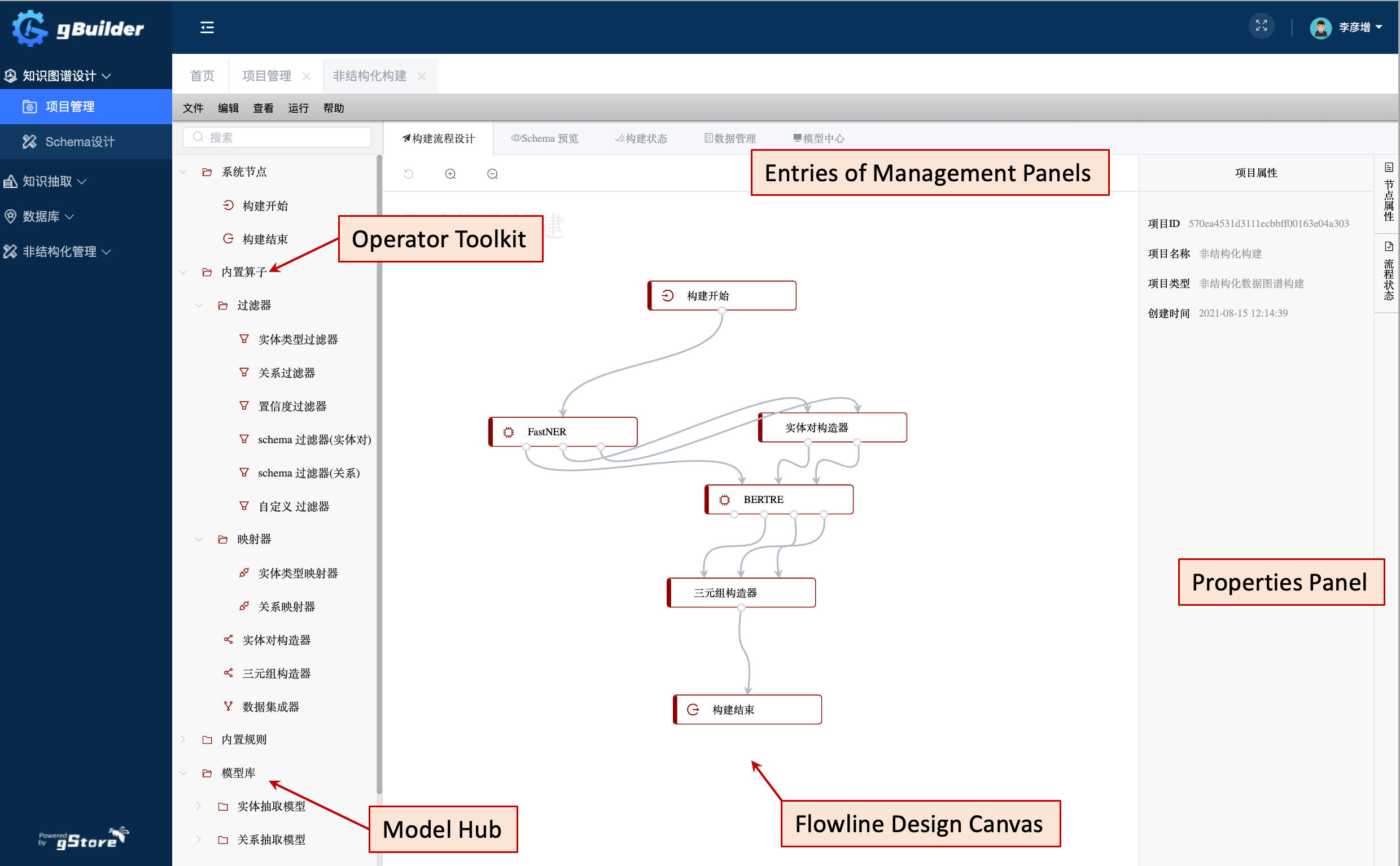}}
\quad
\subfloat[]{\includegraphics[width=3.35in]{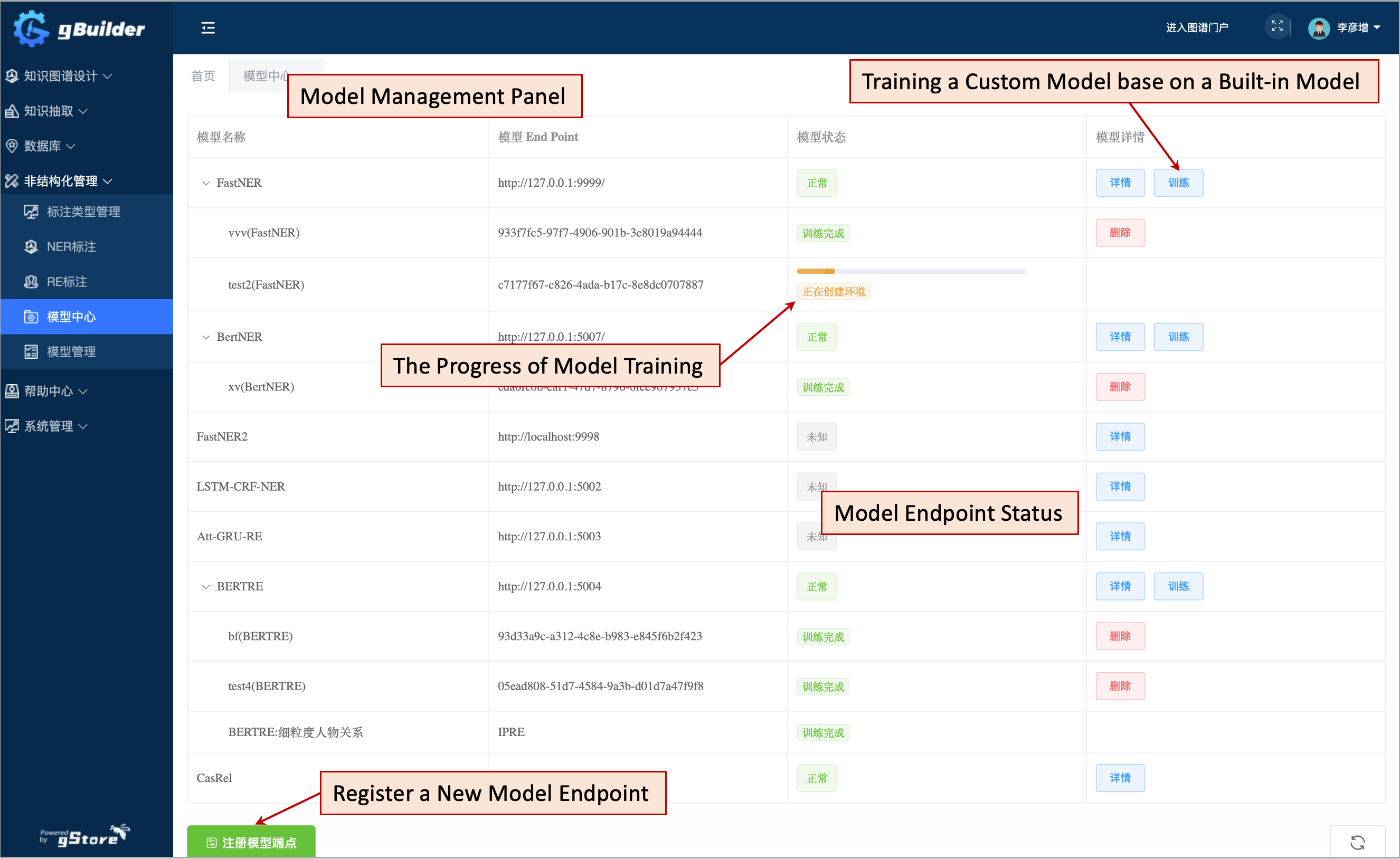}}
\quad
\caption{Partial screenshots from user interfaces. (a)The main interface (including flowline canvas) of gBuilder. (b) The model management interface.}
\label{fig:ui}
\end{figure}

We present partial screenshots from gBuilder in Figure~\ref{fig:ui}.

\begin{figure}[h]
\centering{\includegraphics[width=\linewidth]{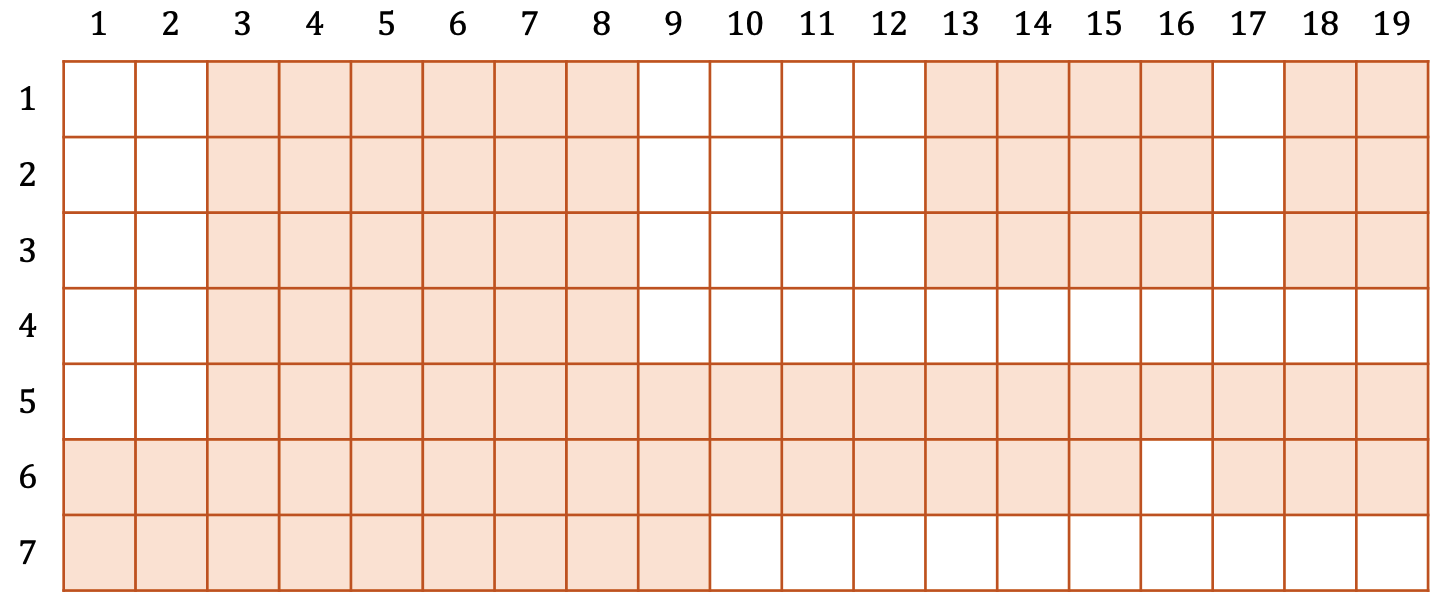}}
	\caption{The "matrix" of built-in models, consists of the built-in models (columns) and the datasets (rows). The colored grid represents that the model has been trained and well-tuned on the corresponding dataset.}
	\label{fig:model-matrix}
\end{figure}

\section{Built-in Models} \label{app:built-in-model}

gBuilder provides a ``matrix'' of built-in models (like \Figure \ref{fig:model-matrix}) with different model structures (as in Table \ref{tab:builtin-models}) and pre-trained in different datasets (as in Table \ref{tab:dataset}). %

\begin{table}[h]
  \centering
  \caption{Part of the built-in IE models. \dag indicates the model is implemented from the original paper author. \ddag indicates the model is implemented from the open-source project. * indicates the model is implemented by ourselves.}
  \label{tab:builtin-models}%
    \begin{tabular}{c|rl|c}
    \hline
    Type  & $id$ & Model & Source \\
    \hline
    \multirow{9}[2]{*}{NER} & 1 & FastNER$_{base}$ & \cite{geng-etal-2021-fasthan}\dag\ddag  \\
		  & 2 & FastNER$_{large}$ & \cite{geng-etal-2021-fasthan}\dag\ddag \\
          & 3 & LSTM-CRF & * \\
          & 4 & BERT$_{base}$ & \ddag \\
          & 5 & BERT$_{large}$ & \ddag \\
          & 6 & RoBERTa & \ddag \\
          & 7 & Electra$_{base}$ & \ddag *  \\
          & 8 & Electra$_{large}$ & \ddag *  \\
          & 9 & FLAT & \cite{li-etal-2020-flat}$\dag\ddag$ \\
    \hline
    \multirow{7}[2]{*}{RE} & 10 & BiGRU-Att & * \\
          & 11 & BiLSTM-Att & * \\
          & 12 & BiRNN-Att & * \\
          & 13 & BERT$_{base}$ & \ddag * \\
          & 14 & BERT$_{large}$ & \ddag * \\
          & 15 & RoBERTa & \ddag * \\
          & 16 & MTB & \cite{baldini-soares-etal-2019-matching}\dag\ddag \\
          & 17 & BERT$_{wwm-ext}$ & \cite{cui-etal-2021-pretrain}\dag\ddag \\
    \hline
    \multirow{2}[2]{*}{JE} & 18 & CasRel$_{BERT}$ & \cite{wei2020CasRel}\dag\ddag \\
          & 19 & TPLinker$_{BERT}$ & \cite{wang-etal-2020-tplinker}\dag\ddag \\
    \hline
    \end{tabular}%
\end{table}

\begin{table}[h]
  \centering
    \caption{Part of employed datasets for well-tuning built-in models. \checkmark indicates the dataset has annotation for the corresponding task.}
  \label{tab:dataset}%
    \begin{tabular}{rlcccc}
    \toprule
    \multicolumn{1}{c}{\multirow{2}{*}{$id$}} & \multicolumn{1}{l}{\multirow{2}{*}{Dataset}} & \multicolumn{1}{c}{\multirow{2}{*}{Language}} & \multicolumn{3}{c}{Annotation} \\
\cmidrule{4-6}       &   &       & NER   & RE    & SPO \\
    \midrule
    1 & NYT\cite{sandhaus2008new}   & English & \checkmark & \checkmark & \checkmark \\
    2 & WebNLG\cite{web_nlg} & English & \checkmark & \checkmark & \checkmark \\
    3 & TACRED\cite{zhang2017tacred} & English & \checkmark & \checkmark & \checkmark \\
    4 & CONLL\cite{tjong-kim-sang-de-meulder-2003-introduction} & English & \checkmark &       &  \\
    5 & DUIE2.0\cite{li2019duie}  & Chinese & \checkmark & \checkmark & \checkmark \\
    6 & IPRE\cite{wang2019ipre}  & Chinese & \checkmark & \checkmark & \checkmark \\
    7 & Resume\cite{zhang-yang-2018-chinese} & Chinese & \checkmark &       &  \\
    \bottomrule
    \end{tabular}%
\end{table}%

\section{Built-in Operators}\label{app:builtin}

gBuilder provides numerous necessary built-in operators for KGC. The most commonly used built-in operators and their descriptions are listed in Table \ref{tab:builtin}.

{
\setlength{\rotFPtop}{0pt plus 1fil}
\setlength{\rotFPbot}{0pt plus 1fil}
\begin{sidewaystable}[ht]
\tiny
  \centering
  \caption{The commonly used built-in operators in gBuilder. (\&) denotes the output is consistent with the input, (*) denotes the operator accepting any input type, (/) denotes there is no input or output.}
    \begin{tabular}{cllllcc}
    \toprule
          & operator name & Description & Input & Output & configurable & programmable \\
    \midrule
    \multirow{5}[10]{*}{filter} & entity type filter & Filter the entities according to configured entity type & entity, entity\_type & \&    & \checkmark &  \\
\cmidrule{2-7}          & relation filter & Filter the relations according to configured relation list & relation\_category & \&    & \checkmark &  \\
\cmidrule{2-7}          & score filter & Filter the entities or relations according to score located in meta info and configured threshold & entity|relation\_category, .meta.score & \&    & \checkmark &  \\
\cmidrule{2-7}          & schema filter & Filter the entities, relations and entity pairs according to designed schema & *     & \&    &       &  \\
\cmidrule{2-7}          & custom filter & Filter based on user-programmed lambda function & *     & \&    &       & \checkmark \\
    \midrule
    \multirow{3}[6]{*}{mapper} & entity type mapper & Map the entity type according to the configured mapping & Class(entity\_type) & \&    & \checkmark &  \\
\cmidrule{2-7}          & relation mapper & Map the relation according to the configured mapping & Relation(relation\_category) & \&    & \checkmark &  \\
\cmidrule{2-7}          & custom mapper & Map the entity type or relation based on user-programmed lambda function & *     & \&    &       & \checkmark \\
    \midrule
    \multirow{4}[8]{*}{integrator} & category ensemble (vote) & Model ensemble for multiple classification-based models based on majority vote & relation\_category & \&    &       &  \\
\cmidrule{2-7}          & category ensemble (score) & Model ensemble for multiple classification-based models based on weighted score sum & relation\_category, .meta.score & \&    & \checkmark &  \\
\cmidrule{2-7}          & chunk ensemble & Integrate chunks like entities obtained from the multiple extraction-based models & entity, entity\_type & \&    &       &  \\
\cmidrule{2-7}          & data integrator & Integrate (merge) output data from different operators & *     & \&    &       &  \\
    \midrule
    \multirow{2}[4]{*}{constructor} & entity pair constructor & Construct the given entities into entity pairs via permutation & entity & entity\_pair &       &  \\
\cmidrule{2-7}          & triple constructor & Construct the given entity pairs and the corresponding relations into triple set & entity\_pair, relation\_category & triple &       &  \\
    \midrule
    \multirow{2}[4]{*}{controller} & start & Denote the entrance of the flowline & /     & sample &       &  \\
\cmidrule{2-7}          & end   & Denote the outlet of the flowline & *     & /     &       &  \\
    \midrule
    \multirow{4}[8]{*}{IE models} & NER   & Extract the entities and the corresponding entity types from the samples & sample & entity, entity\_type &       &  \\
\cmidrule{2-7}          & RE    & Classify the relations of entity pairs according to given samples and entity pairs & sample, entity\_pair & relation\_category &       &  \\
\cmidrule{2-7}          & JE    & Extract entities and simultaneously determine the relationship between entities from given samples & sample & entity\_pair, relation\_category &       &  \\
\cmidrule{2-7}          & AE    & Extract the attributes and attribute values of the specified entity from given samples & sample, entity & attribute, attribute\_value &       &  \\
    \bottomrule
    \end{tabular}%
  \label{tab:builtin}
\end{sidewaystable}%
}

\clearpage
\section{Example Prompts}\label{app:prompts}

In Table~\ref{tab:NER-prompt} and Table~\ref{tab:RE-prompt}, we list some prompts used for extracting named entities and relations. These prompts have been tested in ChatGPT and perform well. 

\begin{table}[h]
    \centering
    \caption{Prompt for extracting named entites guided by give demonstration samples, designed for CONLL-2003 dataset.}
    \resizebox{\linewidth}{!}{
    \begin{tabular}{p{\linewidth}} \toprule
    \textbf{Prompting} \\ \midrule
    \textbf{Task Description:}\newline You are a highly intelligent and accurate news domain named-entity recognition (NER) system. You take passage as input and your task is to recognize and extract specific types of named entities in that given passage and classify into a set of following predefined entity types: 
    [person (PER), location (LOC), organization (ORG), miscellaneous entity (MISC)]. 
    \newline Your output format must be in json form of:  [{``span'': span, ``type'': type}, ...]
    \newline The span must be exactly the same as the original text, including white spaces. \\ \midrule
    \textbf{Demonstrations:}\newline \{``Input": ``EU rejects German call to boycott British lamb.", ``Output": \{"span": [``EU", ``German", ``British"], ``type": [``ORG", ``MISC"]\}\} 
    \newline ... (Some of selected representative samples in CONLL-2003 dataset) \\ \midrule
    \textbf{Instrction:} \newline Following the demonstrations and task description, extract entities and recognize their types, format your output in json strictly. \newline
    \textbf{Input:} \newline
    The European Commission said on Thursday it disagreed with German advice to consumers to shun British lamb until scientists determine whether mad cow disease can be transmitted to sheep. \\
    \bottomrule
    \end{tabular}}
    \label{tab:NER-prompt}
\end{table}

\begin{table}[h]
    \centering
    \caption{Prompt for classifying relationship guided by give demonstration samples, designed for TACRED dataset.}
    \resizebox{\linewidth}{!}{
    \begin{tabular}{p{\linewidth}} \toprule
    \textbf{Prompting} \\ \midrule
    \textbf{Task Description:}\newline Given a sentence, and two entities within the sentence, classify the relationship between the two entities based on the provided sentence. If no relation of interest exists, strictly return ``no\_relation''. All possible relationships and explanations are listed below:
    \newline - per:age : the age of \{e1\} is \{e2\}
    \newline - per:parents : \{e1\}'s parent is \{e2\}
    \newline - per:spouse : \{e1\}'s spouse is \{e2\}
    \newline - per:siblings : \{e1\} is the sibling of \{e2\}
    \newline - per:children : \{e1\}'s children is \{e2\}
    \newline - per:nationality: \{e1\}'s nationality is \{e2\}
    \newline ... (Listing all the verbalized relations in TACRED dataset.)
    \newline - no\_relation : \{e1\} has no known relations to \{e2\}
    \\ \midrule
    \textbf{Demonstrations:}\newline \{``Sentence": ``Boris Collardi, chief executive of Baer, said the deal would strengthen its ties to central and eastern Europe.", ``e1": ``Baer", ``e2": ``Boris Collardi" \} Output: ``org:top\_members"
    \newline ... (Some of selected representative samples in Re-TACRED dataset) \\ \midrule
    \textbf{Instrction:} \newline Following the demonstrations and task description, output the classified relationship according to input sentence and entities.  \newline
    \textbf{Input:} \newline
    \{``Sentence": ``Chavez was also due to meet Sergei Chemezov, who oversees Russian arms export monopoly Rosoboronexport, a Kremin source said Tuesday.", ``e1": ``Rosoboronexport", ``e2'': ``Sergei Chemezov"\} \\
    \bottomrule
    \end{tabular}}
    \label{tab:RE-prompt}
\end{table}

\section{Hardness Analysis}\label{app:hard}

We provide a simple proof for the hardness of Theorem~\ref{thm}.

\begin{proof} Considering with a special case with $\eta = 1$, which means only the $\text{cost}_\text{com}$ would be considered. In this case, the DAG partitioning problem equals the Minimum Makespan Scheduling problem \cite{garey1979computers} with $|V|$ jobs and $k$ machines, which has been proven that the Minimum Makespan Scheduling problem has no polynomial-time approximation algorithm with a finite approximation factor unless $P=NP$ if $k>1$ \cite{10.1145/361011.361064}. Therefore, the theorem holds. 
\end{proof}

\section{Budget-based Minimum Makespan Estimation}\label{app:cost2makespan}

\begin{figure}[!htb]
\centering
\subfloat[]{\includegraphics[width=1.5in]{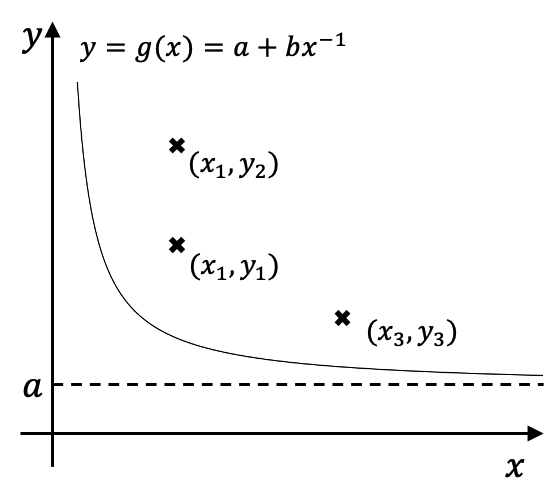}}\label{fig:p2m}
\subfloat[]{\includegraphics[width=1.83in]{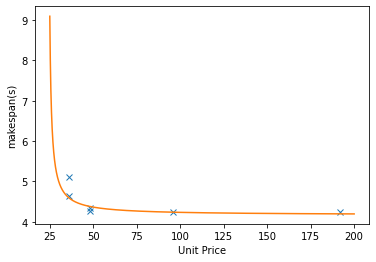}}\label{fig:equ_example_flowline_price}
\caption{(a) The hypothetical dominant function between $P(G)$ to $M(G)$. (b) The visual scatterplot of Table \ref{tab:example_flowline_price} and the boundary curve of \Equation \ref{equ:p2m} under the fitted coefficients ($a = 4.17$, $b=5.15$, $c=23.96$).}
\end{figure}

For convenience, we employ $P(G)$ to denote the unit monetary cost for purchasing VMs with enough capacity to contain the flowline $G$, as $\sum_{i=1}^{i=k}{Price({vm_i})}$ in \Equation \ref{equ:price} in the reality cloud market. Since price is related to the infrastructure cost of cloud vendors (including hardware, electricity, etc.), which are mediately related to the computing power (e.g., FLOPS) and affect the length of makespan, we hypothetically define a function $M(G) = g(P(G))$ to model the relation between price and makespan.
For convenience, we denote it as $y = g(x)$, the function $g$ satisfies the follow properties: 
\begin{itemize}[leftmargin=*,align=left]
    \item Domain of $g$ is $x > 0$ and range of $g$ is $y > 0$;
    \item $\lim _{x \rightarrow +\infty} g(x)= a$, where $a$ is a constant. With a given $G$, suppose ideally that we can purchase a powerful enough VM that meets the resources constraint with monolithic-machine performance, and all the tasks (including IE models and operators) can run on this VM asynchronously and concurrently. Under this condition, the scheduling achieves the optimal performance $cost_{com} \propto M(G)$, the makespan $M(G)$ is equivalent to the critical path of weighted DAG $G$ and is impossible to make a shorter $M(G)$ even with unlimited pricing.
    \item $\lim _{x \rightarrow 0^+} g(x)=+\infty$. Suppose we don't want to spend much money on purchasing VMs, the procurement plan will tend to purchase multiple small-capacity VMs, which will also cause the flowline to be partitioned excessively, resulting in huge communication overhead. In the extreme case, our procurement fund is not enough to meet the resource requirement of the execution of the flowline, which will lead to an infinite makespan $M(G)$.
    \item $g(x_1) \geq g(x_2), x_1 < x_2$. According to common sense, large $P(G)$ would purchase more computing power and reducing makespan $M(G)$ naturally, as $(x_3, y_3)$ and $(x_1, y_1)$ in \Figure \ref{fig:p2m} (a).  
    However, in distributed computing scenarios, we should also consider the situation after partitioning.
Assuming makespan $M(G)$ only depends on CPU and GPU resources: execution of flowline $G$ requires $A$ CPUs and $B$ GPUs. With partitioning $G$ into 2 clusters, suppose one of the sub-clusters requires $a$ CPUs and $b$ GPUs, according to \Equation \ref{equ:price}, $Price(A,B)= Price(a, b) + Price(A-a, B-b)$. 
    Nevertheless, we generally purchase VMs with sufficient GPUs but superfluous CPUs: $Price(A,B) \leq Price(a\prime ,b) + Price(C, B-b)$, where $a\prime \leq a$ and $C \leq A-a$, and it is equivalent to 
    \begin{equation}\label{equ:same_price}
        P(G) \leq P(G^\prime)
    \end{equation}
    For the partitioned $G$, there are generally $M(G) \leq M(G^\prime)$, and the increment is mainly due to the increased communication across VMs. This status is described as $(x_1, y_1)$ and $(x_1, y_2)$ in \Figure \ref{fig:p2m} (a).
\end{itemize}
We define a simple polynomial function
\begin{equation}\label{equ:p2m}
    y = a + b(x - c)^{-1}, (a, b, c, x > 0)
\end{equation}
to model $g$ as \Figure \ref{fig:p2m}, which describes the front of $P(G)$ and $M(G)$. For testing this model in the real-world data, we conduct a verification experiment on qCloud~(Tencent Cloud\footnote{\url{https://intl.cloud.tencent.com}}), which is an IaaS cloud computing platform. The configuration and price (pay-as-you-go billing model) of GPU instance of qCloud's GN10Xp series\footnote{\url{ https://intl.cloud.tencent.com/document/product/560/19701\#GN10Xp}} are listed in Table \ref{tab:qcloud_pricing}.

\begin{table}[ht]
  \centering
  \addtolength{\tabcolsep}{-1pt}
  \caption{Instance model configuration and pricing of qCloud GN10Xp.}
    \label{tab:qcloud_pricing}
    \resizebox{\linewidth}{!}{
    \begin{tabular}{llllll} \toprule
    \multirow{2}{*}{Instance} & GPU & GPU & CPU & Memory & Unit Price \\ 
    &num&memory&cores&size&(per hour)\\
    \midrule
    {2XLARGE40} & {1} & {32} & {10} & {40} & {\textyen 11.98} \\
    {5XLARGE80} & {2} & {64} & {20} & {80} & {\textyen 23.96} \\
    {10XLARGE160} & {4} & {128} & {40} & {160} & {\textyen 47.92} \\
    {20XLARGE320} & {8} & {256} & {80} & {320} & {\textyen 95.84} \\
    \bottomrule
    \end{tabular}%
    }
\end{table}
We employ cloud instances from the above products for estimating the relationship between makespan and pricing. The pre-defined flowline as \Figure \ref{fig:flowline1} is introduced as the workload, and the scheduler algorithm proposed in this paper is applied for scheduling. We conducted the experiment 5 times and reported the average value. The results are shown in Table \ref{tab:example_flowline_price}.
\begin{table}[htbp]
  \centering\small
  \addtolength{\tabcolsep}{-1pt}
  \caption{Price and makespan for executing example Flowline (as \Figure \ref{fig:flowline1}) in real-world. The bolded ``\textbf{($\cdot$)X}'' denotes the corresponding instance model ``($\cdot$)XLARGE'' in Table \ref{tab:qcloud_pricing}.} 
  \label{tab:example_flowline_price}%
  \resizebox{\linewidth}{!}{
    \begin{tabular}{llrrr}\toprule
    No. & VM group & Makespan(s) & Total Price (\textyen) & Unit Price (\textyen)  \\\midrule
    0 & \textbf{2X} $\times$ 2 & $\infty$ & - & 23.96 \\
    1 & \textbf{2X} $\times$ 3 & 5.10   & 0.050 & 35.94\\
    2 & \textbf{5X} $\times$ 1 + \textbf{2X} $\times$ 1 & 4.65  & 0.046 & 35.94\\
    3 & \textbf{5X} $\times$ 2 & 4.34  & 0.058 & 47.92 \\
    4 & \textbf{10X} $\times$ 1 & 4.26  & 0.056 & 47.92 \\
    5 & \textbf{20X} $\times$ 1 & 4.25  & 0.113 & 95.84 \\
    6 & \textbf{20X} $\times$ 2 & 4.25  & 0.226 & 191.68 \\
    \bottomrule
    \end{tabular}%
    }
\end{table}
The No.0 procurement plan ($\#GPU < |V^m|$) does not satisfy the requirement of resources that would commit an unacceptable makespan length. Using non-linear least squares to fit the boundary curve of scatters would produce $a = 4.17$, $b=5.15$ and $c=23.96$, as \Figure \ref{fig:equ_example_flowline_price} shows.

Substituting the hypothesis polynomial function relationship between price and makespan into the objective function, it can be rewritten as:
\begin{equation}\label{equ:p2j}
J = f(x) = \eta g(x) + (1-\eta) x \cdot g(x)
\end{equation}
which is a convex function with a global minimum at $x_0 = \sqrt{\frac{b}{a} \frac{\eta}{1 - \eta}} + c$, and our objective target is to approach the minimum as possible (Since the procurement plan of VMs is discrete, it is unpractical to close $x_0$ continuously). The value of constants $a$ and $b$ depend on the cloud vendors, and the weight of $\eta$ is determined by user.

\end{document}